\documentclass[compsoc,conference,a4paper,10pt,times]{IEEEtran}
\IEEEoverridecommandlockouts
\usepackage{cite}
\usepackage{amsmath,amssymb,amsfonts}
\usepackage{algorithmic}
\usepackage{graphicx}
\usepackage{textcomp}
\usepackage{bmpsize}
\usepackage{xcolor}
\usepackage{lipsum}
\usepackage[colorlinks=true,urlcolor=black]{hyperref}
\def\BibTeX{{\rm B\kern-.05em{\sc i\kern-.025em b}\kern-.08em
    T\kern-.1667em\lower.7ex\hbox{E}\kern-.125emX}}

\usepackage{epsfig,amsmath,amsfonts,epsfig,multirow,makecell,caption,soul,csquotes,color,wrapfig,subcaption,mathtools,bm,spverbatim,booktabs,tcolorbox,diagbox,todonotes}
\usepackage[e]{esvect}

\def\tablename{Table}
\def\figurename{Figure}

\usepackage{caption}
\makeatletter
\newif\if@restonecol
\makeatother

\usepackage[boxed, ruled, vlined, linesnumbered]{algorithm2e}
\SetKwRepeat{Do}{do}{while}

\newenvironment{changemargin}[2]{\begin{list}{}{
	\setlength{\topsep}{0pt}\setlength{\leftmargin}{0pt}
	\setlength{\rightmargin}{0pt}
	\setlength{\listparindent}{\parindent}
	\setlength{\itemindent}{\parindent}
	\setlength{\parsep}{0pt plus 1pt}
	\addtolength{\leftmargin}{#1}\addtolength{\rightmargin}{#2}
	}\item}
	{\end{list}}

\newcommand{\ssup}[2]{{#1}^{\scaleobj{0.8}{#2}}}
\newcommand{\ssub}[2]{{#1}_{\scaleobj{0.8}{#2}}}

\usepackage[first=0,last=9]{lcg}
\usepackage{colortbl}
\definecolor{Gray}{gray}{0.8}
\colorlet{Red}{red!10!white}
\colorlet{Blue}{blue!10!white}

\usepackage{hyperref}

\newcommand{\msec}[1]{\S\,\ref{#1}}
\newcommand{\mref}[1]{\,\ref{#1}}
\newcommand{\meq}[1]{Eqn.\,\ref{#1}}
\newcommand{\mcite}[1]{\cite{#1}}

\newcommand{\meg}{\textit{e.g.}\xspace}
\newcommand{\mie}{\textit{i.e.}\xspace}
\newcommand{\mcf}{\textit{cf}.\xspace}
\newcommand{\mcounter}[1]{(\textit{#1})}

\newtcolorbox{mtbox}[1]{left=0.25mm, right=0.25mm, top=0.25mm, bottom=0.25mm, sharp corners, colframe=blue!50!black, boxrule=0.5pt, title={#1}, fonttitle=\bfseries, coltitle=blue!50!black, attach title to upper={\ --\ }}

\usepackage{scalerel}[2016/12/29]

\makeatletter
\providecommand{\leadsfrom}{%
  \mathrel{\mathpalette\reflect@squig\relax}%
}
\newcommand{\reflect@squig}[2]{%
  \reflectbox{$\m@th#1\leadsto$}%
}
\makeatother

\newcommand{\bx}{x}
\newcommand{\ax}{x}
\newcommand{\ux}{\ssup{x}{*}}

\newcommand{\ay}{t}

\newcommand{\sg}{\ssup{g}{*}}
\newcommand{\af}{f}
\newcommand{\uf}{\ssup{f}{*}}

\newcommand{\dnn}{DNN\xspace}
\newcommand{\dnns}{DNNs\xspace}
\newcommand{\ml}{ML\xspace}

\newcommand{\cifar}{{\small CIFAR10}\xspace}
\newcommand{\ncifar}{{\small CIFAR100}\xspace}
\newcommand{\vggface}{{\small VGGF}ace{\small 2}\xspace}
\newcommand{\gtsrb}{{\small GTSRB}\xspace}
\newcommand{\imgnet}{{\small I}mage{\small N}et\xspace}

\def\eqref#1{equation~\ref{#1}}

\def\1{\bm{1}}

\DeclareMathAlphabet{\mathsfit}{\encodingdefault}{\sfdefault}{m}{sl}
\SetMathAlphabet{\mathsfit}{bold}{\encodingdefault}{\sfdefault}{bx}{n}

\def\gA{{\mathcal{A}}}

\def\gD{{\mathcal{D}}}

\def\gF{{\mathcal{F}}}

\def\gN{{\mathcal{N}}}

\def\gR{{\mathcal{R}}}

\def\gT{{\mathcal{T}}}

\def\sE{{\mathbb{E}}}

\def\sR{{\mathbb{R}}}
\def\sS{{\mathbb{S}}}

\usepackage{caption}
\usepackage{subcaption}
\usepackage{enumitem}
\usepackage{bbm}
\usepackage{svg}
\usepackage[nointegrals]{wasysym}

\renewcommand{\thetable}{\arabic{table}}

\newcommand{\system}{{\sc \small TrojanZoo}\xspace}

\newcommand{\minitab}[2][l]{\begin{tabular}{#1}#2\end{tabular}}

\newcommand{\bn}{{\sc Bn}\xspace}
\newcommand{\tnn}{{\sc Tnn}\xspace}
\newcommand{\tb}{{\sc Tb}\xspace}
\newcommand{\lb}{{\sc Lb}\xspace}
\newcommand{\esb}{{\sc Esb}\xspace}
\newcommand{\rfb}{{\sc Rb}\xspace}

\newcommand{\abe}{{\sc Abe}\xspace}

\newcommand{\imc}{{\sc Imc}\xspace}

\newcommand{\asr}{{\em \small ASR}\xspace}
\newcommand{\cad}{{\em \small CAD}\xspace}
\newcommand{\tmc}{{\em \small TMC}\xspace}
\newcommand{\ccc}{{\em \small CCC}\xspace}
\newcommand{\auc}{{\em \small AUC}\xspace}

\newcommand{\acc}{{\em \small ACC}\xspace}
\newcommand{\ard}{{\em \small ARD}\xspace}
\newcommand{\tpr}{{\em \small TPR}\xspace}
\newcommand{\fpr}{{\em \small FPR}\xspace}
\newcommand{\mln}{{\em \small MLN}\xspace}
\newcommand{\mad}{{\em \small MAD}\xspace}

\newcommand{\nsr}{{\em \small NSR}\xspace}
\newcommand{\art}{{\em \small ART}\xspace}
\newcommand{\mjs}{{\em \small MJS}\xspace}

\newcommand{\anidx}{{\em \small AIV}\xspace}

\newcommand{\nc}{{\sc Nc}\xspace}
\newcommand{\ninspect}{{\sc Ni}\xspace}
\newcommand{\du}{{\sc Du}\xspace}
\newcommand{\abs}{{\sc Abs}\xspace}
\newcommand{\tabor}{{\sc Tabor}\xspace}
\newcommand{\strip}{{\sc Strip}\xspace}
\newcommand{\ac}{{\sc Ac}\xspace}
\newcommand{\di}{{\sc Di}\xspace}
\newcommand{\fp}{{\sc Fp}\xspace}
\newcommand{\at}{{\sc Ar}\xspace}
\newcommand{\rands}{{\sc Rs}\xspace}

\newcommand{\neo}{{\sc Neo}\xspace}
\newcommand{\mss}{{\sc Ss}\xspace}

\newcommand{\mmp}{{\sc Mp}\xspace}

\begin{document}

\title{TrojanZoo: Towards Unified, Holistic, and Practical Evaluation of Neural Backdoors}

\author{
\IEEEauthorblockN{Ren Pang\IEEEauthorrefmark{1}
Zheng Zhang\IEEEauthorrefmark{1}
Xiangshan Gao\IEEEauthorrefmark{2}
Zhaohan Xi\IEEEauthorrefmark{1}}

\IEEEauthorblockN{Shouling Ji\IEEEauthorrefmark{2}
Peng Cheng\IEEEauthorrefmark{2}
Xiapu Luo\IEEEauthorrefmark{3}
Ting Wang\IEEEauthorrefmark{1}}

\IEEEauthorblockA{\IEEEauthorrefmark{1} Pennsylvania State University, \{rbp5354, zxz147, zxx5113, ting\}@psu.edu}
\IEEEauthorblockA{\IEEEauthorrefmark{2} Zhejiang University, \{corazju, sji, lunar\_heart\}@zju.edu.cn}
\IEEEauthorblockA{\IEEEauthorrefmark{3} Hong Kong Polytechnic University, csxluo@comp.polyu.edu.hk}
}

\maketitle
\thispagestyle{plain}
\pagestyle{plain}
\pagenumbering{arabic}

\begin{abstract}
Neural backdoors represent one primary threat to the security of deep learning systems. The intensive research has produced a plethora of backdoor attacks/defenses, resulting in a constant arms race. However, due to the lack of evaluation benchmarks, many critical questions remain under-explored: (i) what are the strengths and limitations of different attacks/defenses? (ii) what are the best practices to operate them? and (iii) how can the existing attacks/defenses be further improved?

To bridge this gap, we design and implement TROJANZOO, the first open-source platform for evaluating neural backdoor attacks/defenses in a unified, holistic, and practical manner. Thus far, focusing on the computer vision domain, it has incorporated 8 representative attacks, 14 state-of-the-art defenses, 6 attack performance metrics, 10 defense utility metrics, as well as rich tools for in-depth analysis of the attack-defense interactions. Leveraging TROJANZOO, we conduct a systematic study on the existing attacks/defenses, unveiling their complex design spectrum: both manifest intricate trade-offs among multiple desiderata (e.g., the effectiveness, evasiveness, and transferability of attacks). We further explore improving the existing attacks/defenses, leading to a number of interesting findings: (i) one-pixel triggers often suffice; (ii) training from scratch often outperforms perturbing benign models to craft trojan models; (iii) optimizing triggers and trojan models jointly greatly improves both attack effectiveness and evasiveness; (iv) individual defenses can often be evaded by adaptive attacks; and (v) exploiting model interpretability significantly improves defense robustness. 
We envision that TROJANZOO will serve as a valuable platform to facilitate future research on neural backdoors.

\end{abstract}


\section{Introduction}

Today's deep learning (DL) systems are large, complex software artifacts. With the increasing system complexity and training cost, it becomes not only tempting but also necessary to exploit pre-trained deep neural networks (DNNs) in building DL systems. It was estimated that as of 2016, over 13.7\% of DL-related repositories on GitHub re-use at least one pre-trained DNN\mcite{Ji:2018:ccsa}. On the upside, this ``plug-and-play'' paradigm greatly simplifies the development cycles\mcite{Sculley:2015:nips}. On the downside, as most pre-trained DNNs are contributed by untrusted third parties\mcite{modelzoo}, their lack of standardization or regulation entails profound security implications.

In particular, pre-trained DNNs can be exploited to launch {\em neural backdoor} attacks\mcite{badnet,trojannn,imc}, one primary threat to the security of DL systems. In such attacks, a maliciously crafted DNN (``trojan model'') forces its host system to misbehave once certain pre-defined conditions (``triggers'') are met but to function normally otherwise, which can result in consequential damages in security-sensitive domains\mcite{Versprille:2015:news,Cooper:2014:news,Biggio:2012:spr}.

Motivated by this, intensive research has led to a plethora of attacks that craft trojan model via exploiting properties such as neural activation patterns\mcite{badnet,targeted-backdoor,trojannn,invisible-backdoor,Suciu:2018:sec,latent-backdoor} and defenses that mitigate trojan models during  inspection\mcite{neural-cleanse,abs,deep-inspect,tabor,neuron-inspect,fine-pruning} or detect trigger inputs at inference\mcite{strip,active-clustering,randomized-smoothing,neo}. With the rapid development of new attacks/defenses, a number of open questions have emerged:  RQ$_1$ -- {\em What are the strengths and limitations of different attacks/defenses?} RQ$_2$ -- {\em What are the best practices (\meg, optimization strategies) to operate them?} RQ$_3$ -- {\em How can the existing backdoor attacks/defenses be further improved?}

\vspace{2pt}

Despite their importance for understanding and mitigating the vulnerabilities incurred by neural backdoors, these questions are largely under-explored due to the following challenges. 

\vspace{1pt}
\underline{Non-holistic evaluations} -- Most studies conduct the evaluation with a fairly limited set of attacks/defenses, resulting in incomplete assessment. For instance, it is unknown whether {\strip}\mcite{strip} is effective against the newer {\abe} attack\mcite{adv-backdoor}. Further, the evaluation often uses simple, macro-level metrics, failing to comprehensively characterize given attacks/defenses. For instance, most studies use attack success rate (\asr) and clean accuracy drop (\cad) to assess attack performance, which is insufficient to describe the attack's ability of trading between these two metrics.  

\vspace{1pt}
\underline{Non-unified platforms} -- Due to the lack of unified benchmarks, different attacks/defenses are often evaluated under inconsistent settings, leading to non-comparable conclusions. For instance, {\tnn}\mcite{trojannn} and {\lb}\mcite{latent-backdoor} are evaluated with distinct trigger definitions (\mie,\,shape, size, and transparency), datasets, and DNNs, making it difficult to directly compare their assessment. 

\vspace{1pt}
\underline{Non-adaptive attacks} --  The evaluation of the existing defense\mcite{fine-pruning,neural-cleanse,strip,tabor} often assume static, non-adaptive attacks, without fully accounting for the adversary's possible countermeasures, which however is critical for modeling the adversary's optimal strategies and assessing the attack vulnerabilities in realistic settings.

\subsection*{Our Work} To this end, we design, implement, and evaluate \system, an open-source platform for assessing neural backdoor attacks/defenses in a unified, holistic, and practical manner. Note that while it is extensible to other domains (\meg, NLP), currently, \system focuses on the image classification task in the computer vision domain. Our contributions are summarized in three major aspects:

\vspace{2pt}
{\bf Platform --} To our best knowledge, \system represents the first open-source platform specifically designed for evaluating neural backdoor attacks/defenses. At the moment of writing (02/06/2022), focusing on the computer vision domain, \system has incorporated 8 representative attacks, 14 state-of-the-art defenses, 6 attack performance metrics, 10 defense utility metrics, as well as a benchmark suite of 5 DNN models, 5 downstream models, and 6 datasets. Further, \system implements a rich set of tools for in-depth analysis of the attack-defense interactions, including measuring feature-space similarity, tracing neural activation patterns, and comparing attribution maps.

\vspace{2pt}
{\bf Measurement --} Leveraging \system, we conduct a systematic study of the existing attacks/defenses, unveiling the complex design spectrum for the adversary and the defender. Different attacks manifest delicate trade-offs among effectiveness, evasiveness, and transferability. For instance, weaker attacks (\mie, lower \asr) tend to show higher transferability. Meanwhile, different defenses demonstrate trade-offs among robustness, utility-preservation, and detection accuracy. For instance, while effective against a variety of attacks, model sanitization\mcite{madry:iclr:2018,fine-pruning} also incur a significant accuracy drop. These observations indicate the importance of using comprehensive metrics to evaluate neural backdoor attacks/defenses, and suggest the optimal practices of applying them under given settings.

\vspace{2pt}
{\bf Exploration --} We further explore improving existing attacks/defenses, leading to a number of previously unknown findings including \mcounter{i} one-pixel triggers often suffice (over 95\% \asr); \mcounter{ii} training from scratch often outperforms perturbing benign models to forge trojan models; \mcounter{iii} leveraging DNN architectures (\meg,\,skip connects) in optimizing trojan models improves the attack effectiveness; \mcounter{iv} most individual defenses are vulnerable to adaptive attacks; and \mcounter{v} exploiting model interpretability significantly improves defense robustness. We envision that the \system platform and our findings will facilitate future research on neural backdoors and shed light on designing and building DL systems in a more secure and informative manner.\footnote{All the data, models, and code used in the paper are released at:\\ \url{https://github.com/ain-soph/trojanzoo}.}

\subsection*{Roadmap}

The remainder of the paper proceeds as follows. \msec{sec:background} introduces fundamental concepts and assumptions; \msec{sec:platform} details the design and implementation of \system and systemizes existing attacks/defenses; equipped with \system, \msec{sec:evaluation} conducts a systematic evaluation of existing attacks/defenses; \msec{sec:exploration} explores their further improvement; \msec{sec:discussion} discusses the limitations of \system and points to future directions; the paper is concluded in \msec{sec:conclusion}.

\begin{figure}[!t]
\centering
\includegraphics[width=80mm]{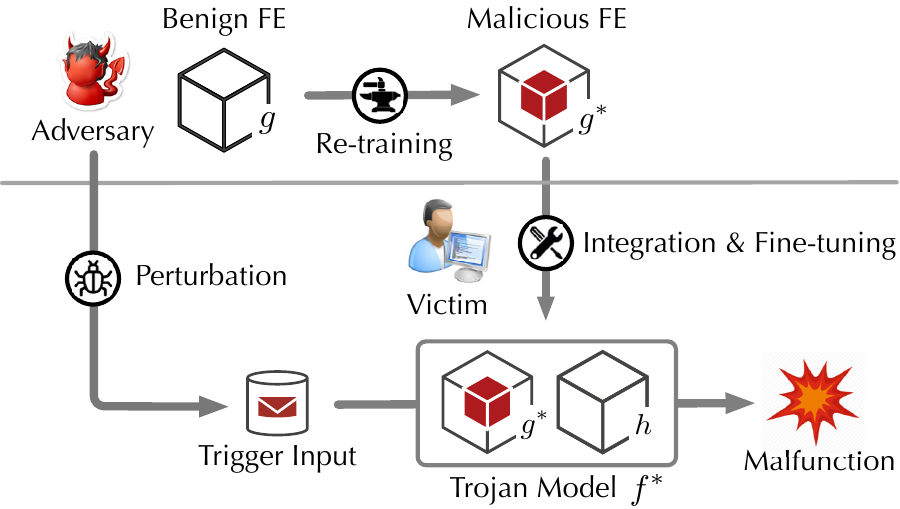}
\caption{Illustration of neural backdoor attacks.}
\label{fig:attack}
\end{figure}

\section{Related Work}

Some recent studies have surveyed neural backdoor attacks/defenses (\meg, \mcite{survey1}); yet, none of them provides benchmark implementation or empirical evaluation to explore their strengths/limitations. Compared with the rich collection of platforms for adversarial attacks/defenses (\meg, {\sc CleverHans}\mcite{cleverhans}, {\sc DeepSec}\mcite{Ling:2019:sp}, and {\sc AdvBox}\mcite{advbox}), only few platforms currently support evaluating neural backdoors. For instance, {\sc Art}\mcite{ibm-art} integrates 3 attacks and 3 defenses.

In comparison, \system differs in major aspects: \mcounter{i} to our best knowledge, it features the most comprehensive library of attacks/defenses; \mcounter{ii} it regards the evaluation metrics as a first-class citizen and implements 6 attack performance metrics and 10 defense utility metrics, which holistically assess given attacks/defenses; \mcounter{iii} besides reference implementation, it also provides rich utility tools for in-depth analysis of attack-defense interactions, such as measuring feature-space similarity, tracing neural activation patterns, and comparing attribution maps.
r
The work closest to ours is perhaps {\sc TrojAI}\mcite{trojai}, which is a contest platform for model-inspection defenses against neural backdoors. While compared with \system, {\sc TrojAI} provides a much larger pool of trojan models (over 10K) across different modalities (\meg, vision and NLP), \system departs from {\sc TrojAI} in majors aspects and offers its unique value. \mcounter{i} Given its contest-like setting, {\sc TrojAI} is a closed platform focusing on evaluating model-inspection defenses (\mie, detecting trojan models) against fixed attacks, while \system is an open platform that provides extensible datasets, models, attacks, and defenses. Thus, \system may serve the needs ranging from conducting comparative studies of existing attacks/defenses to exploring and evaluating new attacks/defenses. \mcounter{ii} While {\sc TrojAI} focuses on model-inspection defenses,  \system integrates four major defense categories. \mcounter{iii} In {\sc TrojAI}, for its purpose, the concrete attacks behind the trojan models are unknown, which makes it challenging to assess the strengths/limitations of given defenses with respect
to different attacks, while in \system one may directly evaluate such interactions. \mcounter{iv} As the attacks are fixed in {\sc TrojAI}, one may not evaluate adaptive attacks. \mcounter{v} The main metric used in {\sc TrojAI} is the accuracy that defenses successfully detect trojan models, while \system provides a much richer set of metrics to characterize attacks/defenses.

\section{Fundamentals}
\label{sec:background}


We first introduce fundamental concepts and assumptions used throughout the paper.
The important notations are summarized in Table\mref{tab:symbol}.

\begin{table}[!ht]{\footnotesize
    \centering
       \renewcommand{\arraystretch}{1.2}
    \begin{tabular}{r|l}
    {Notation} & {Definition}\\
    \hline
    \hline
    $\gA, \gD$ & attack, defense\\
    $x, \ux$ & clean input, trigger input \\
    $x_i$ & $i$-th dimension of $x$\\
    $r$ & trigger \\
    $m$ & mask ($\alpha$ for each pixel) \\
    $f, \uf$ & benign model, trojan model\\
    $f_{\text{feat}}$ & upstream feature extractor\\
    $g, \sg$ & downstream classifier, surrogate classifier\\
    $\ay$ & adversary's target class \\
    $\mathcal{T}$ & reference set\\
    $\gR_\epsilon, \gF_\delta$ & trigger, model feasible sets
    \end{tabular}
    \caption{Symbols and notations. \label{tab:symbol}}}
\end{table}

\subsection{Preliminaries}

{\bf Deep neural networks (DNNs) --} 
Deep neural networks (\dnns) represent a class of \ml models to learn high-level abstractions of complex data. We assume a predictive setting, in which a \dnn $\ssub{f}{\theta}$ (parameterized by $\theta$) encodes a function $\ssub{f}{\theta}: \ssup{\sR}{n} \rightarrow \ssup{\sS}{m}$, where $n$ and $m$ denote the input dimensionality and the number of classes. Given input $x$, $f(x)$ is a probability vector (simplex) over $m$ classes.

\vspace{2pt}
{\bf Pre-trained DNNs --} Today, it becomes not only tempting but also necessary to reuse pre-trained models in domains in which data labeling or model training is expensive\mcite{transfer-learning}. Under the transfer learning setting, as shown in Figure\mref{fig:attack}, the feature extractor (FE) $g$ of a pre-trained model is often reused and composed with a classifier $h$ to form an end-to-end model $f$. As the data used to train $g$ may differ from the downstream task, it is often necessary to fine-tune $f = h \circ g$ in a supervised manner. One may opt to perform full-tuning to train both $g$ and $h$ or partial-tuning to train $h$ only with $g$ fixed\mcite{Ji:2018:ccsa}.

\begin{figure*}[!ht]
    \centering
    \includegraphics[width=170mm]{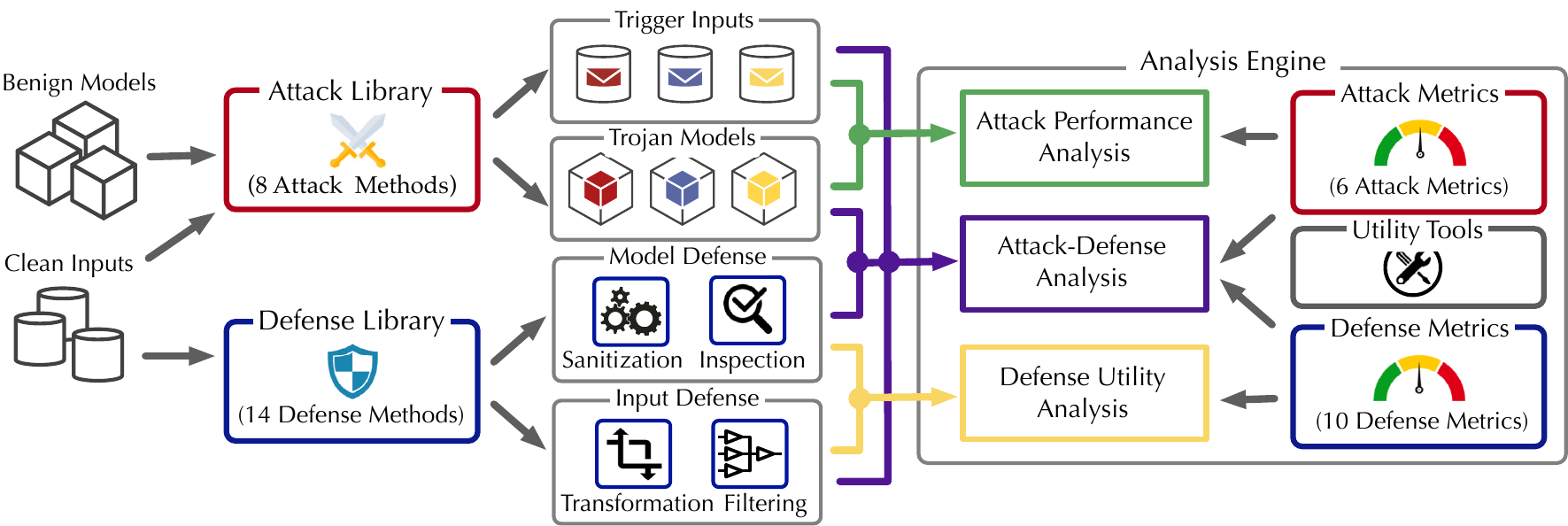}
    \caption{Overall system design of \system.}
    \label{fig:framework}
\end{figure*}

\vspace{2pt}
{\bf Neural backdoor attacks --} With the increasing use of DNN models in security-sensitive domains, the adversary is strongly incentivized to forge malicious FEs as attack vectors and lure victim users to re-use them during system development\mcite{badnet}. Specifically, through a malicious FE, the backdoor attack infects the target model with malicious functions desired by the adversary, which are activated once pre-defined conditions (``triggers'') are present. 
We refer to such infected models as ``trojan models''. Typically, a trojan model reacts to trigger-embedded inputs (\meg,\,images with specific watermarks) in a highly predictable manner (\meg,\,misclassified to a target class) but functions normally otherwise.

\subsection{Specifics}

\noindent
\hspace{\parindent}
{\bf Trigger mixing operator --} For given trigger $r$, the operator $\oplus$ mixes a clean input $\bx \in \ssup{\mathbb{R}}{n}$ with $r$ to generate a trigger input $\bx \oplus r$. Typically, $r$ comprises three parts: ({\em i}) mask $m \in \ssup{\{0, 1\}}{n}$ specifies where $r$ is applied (\mie, $\bx$'s $i$-th feature $\ssub{\bx}{i}$ is retained if $\ssub{m}{i}$ is on and mixed with $r$ otherwise); ({\em ii}) transparency $\alpha \in [0, 1]$ specifies the mixing weight; and ({\em iii}) pattern $p(\bx) \in \ssup{\mathbb{R}}{n}$ specifies $r$'s color intensity, which can be a constant, randomly drawn from a distribution (\meg, by perturbing a template), or dependent on $\bx$\mcite{adaptive-trigger}. Formally, the trigger embedding operator is defined as:
\begin{align}
\label{eq:trigger}
\bx \oplus r = (1-m) \odot [(1 - \alpha) x + \alpha p(\bx) ] + m \odot \bx
\end{align}
where $\odot$ denotes element-wise multiplication.

\vspace{2pt}
{\bf Attack objectives --} The trojan model satisfies that with high probability, \mcounter{i} trigger inputs are classified to the target class desired by the adversary and \mcounter{ii} clean input are still correctly classified.
Formally, the adversary forges the malicious FE by optimizing the following objective:
\begin{align}
\label{eq:opt}
\min_{r \in \gR, \theta} \sE_{(x,y) \in \mathcal{T}}\left[ \ell( \ssub{f}{\theta}(x \oplus r), \ay)  + \lambda \ell( \ssub{f}{\theta}(x), y) \right]
\end{align}
where $\gT$ represents the training set, $t$ denotes the target class, and trigger $r$ is selected from the feasible set $\gR$ (which constrains $r$'s shape, transparency, and/or pattern). Intuitively, the first and second terms describe \mcounter{i} and \mcounter{ii}, respectively, and the hyper-parameter $\lambda$ balances the two objectives.

\vspace{2pt}
{\bf Adversary's knowledge --} If the downstream classifier $h$ is known to the adversary, $f$ shares the same architecture with the model $h \circ g$ used by the victim; otherwise, the adversary may resort to a surrogate classifier $\ssup{h}{*}$ (\mie, $ \ssup{h}{*} \circ g$) or re-define the loss $\ell(f(x \oplus r), \ay)$ in terms of latent representations\mcite{latent-backdoor,imc} as $\Delta (g(x\oplus r), \phi_\ay)$, that is, the difference(\meg, MSE loss) between $g(x\oplus r)$ and $\ssub{\phi}{\ay}$, where $\ssub{\phi}{\ay}$ is the average latent representation of class $\ay$. 

\vspace{2pt}
{\bf Malicious FE training --}
To optimize \meq{eq:opt}, one may perturb a benign FE\mcite{Suciu:2018:sec,trojannn} or train the malicious FE from scratch (details in \msec{sec:exploration}).
To satisfy the trigger constraint, $r$ can be fixed\mcite{badnet}, partially defined\mcite{trojannn} (\meg,\,with its mask fixed), or optimized with $f$ jointly\mcite{imc}. 

\section{Platform}
\label{sec:platform}

As illustrated in Figure\mref{fig:framework}, \system comprises three major components: \mcounter{i} the attack library integrates a set of representative attacks that, for given benign models and clean inputs, are able to generate trojan models and trigger inputs; \mcounter{ii} the defense library integrates a set of state-of-the-art defenses that are able to provide model- and input-level protection against trojan models and trigger inputs;  
and \mcounter{iii} the analysis engine, equipped with attack performance metrics, defense utility metrics, and feature-rich utility tools, is able to conduct unified and holistic evaluation across different attacks/defenses.

In its current implementation, \system has incorporated 8 attacks, 14 defenses, 6 attack performance metrics, and 10 defense utility metrics, which we systematize as follows.

\subsection{Attack Library}

While neural backdoor attacks can be characterized from a number of aspects, here we focus on 4 key design choices by the adversary that directly impact attack performance. Table\mref{tab:attack_summary} summarizes the representative neural backdoor attacks currently implemented in \system, which are characterized along the above 4 dimensions. More specifically,

\begin{table}[!ht]{\footnotesize
\centering
\renewcommand{\arraystretch}{1.2}
\setlength{\tabcolsep}{5pt}
\begin{tabular}{r|c|c|c|c}
\multirow{2}{*}{Attack} & {Architecture} & {Trigger}  & { Fine-tuning} & {Defense}
\\
& {Modifiability} & {Optimizability} &  {Survivability} & {Adaptivity}\\
\hline
\hline
{\bn}\mcite{badnet} & $\Circle$ & $\Circle$ & $\Circle$ & $\Circle$ \\
{\esb}\mcite{embarassingly-simple-backdoor} & $\CIRCLE$ & $\Circle$ & $\Circle$ & $\Circle$ \\
{\tnn}\mcite{trojannn} & $\Circle$ & $\LEFTcircle$ & $\Circle$ & $\Circle$ \\
{\rfb}\mcite{reflection-backdoor} & $\Circle$ & $\LEFTcircle$ & $\Circle$ & $\Circle$ \\
{\tb}\mcite{targeted-backdoor} & $\Circle$ & $\LEFTcircle$ & $\Circle$ & $\Circle$ \\
{\lb}\mcite{latent-backdoor} & $\Circle$ & $\Circle$ & $\CIRCLE$ & $\Circle$ \\
{\abe}\mcite{adv-backdoor} & $\Circle$ & $\Circle$ & $\Circle$ & $\CIRCLE$ \\
{\imc}\mcite{imc} & $\Circle$ & $\CIRCLE$  & $\CIRCLE$ & $\CIRCLE$ \\
\end{tabular}
\caption{Summary of representative neural backdoor attacks currently implemented in \system ($\CIRCLE$ -- full optimization, $\LEFTcircle$ -- partial optimization, $\Circle$ -- no optimization)}
\label{tab:attack_summary}
}
\end{table}

\vspace{2pt}
{\bf Non-optimization --} The attack simply solves \meq{eq:opt} under pre-defined triggers (\mie, shape, transparency, and pattern) without optimization for other desiderata.

-- BadNet ({\bn})\mcite{badnet}, as the representative, pre-defines trigger $r$, generates trigger inputs $\{ (x\oplus r, t)\}$, and crafts the trojan model $\uf$ by re-training a benign model $f$ with such data.

\vspace{2pt}
{\bf Architecture modifiability --} whether the attack is able to change the DNN architecture. Being allowed to modify both the architecture and the parameters enables a larger attack spectrum, but also renders the trojan model more susceptible to certain defenses (\meg,\,model specification checking).

-- Embarrassingly-Simple-Backdoor (\esb)\mcite{embarassingly-simple-backdoor}, as the representative, modifies $f$'s architecture by adding a module which overwrites the prediction as $\ay$ if $r$ is recognized. Without disturbing $f$'s original configuration, $\uf$ retains $f$'s predictive power on clean inputs.

\vspace{2pt}
{\bf Trigger optimizability --} whether the attack uses a fixed, pre-defined trigger or optimizes it during crafting the trojan model. Trigger optimization often leads to stronger attacks with respect to given desiderata (\meg,\,trigger stealthiness).

-- TrojanNN (\tnn)\mcite{trojannn} fixes $r$'s shape and position, optimizes its pattern to activate neurons rarely activated by clean inputs in pre-processing, and then forges $\uf$ by re-training $f$ in a manner similar to \bn.

-- Reflection-Backdoor (\rfb)\mcite{reflection-backdoor} optimizes trigger stealthiness by defining $r$ as the physical reflection of a clean image $\ssup{x}{r}$ (selected from a pool): $r = \ssup{x}{r} \otimes k$, where $k$ is a convolution kernel, and $\otimes$ is the convolution operator.

-- Targeted-Backdoor (\tb)\mcite{targeted-backdoor} randomly generates $r$'s position in training, which makes $\uf$ effective regardless of $r$'s position and allows the adversary to optimize $r$'s stealthiness by placing it at the most plausible position (\meg,\,an eyewear watermark over eyes).

\vspace{2pt}
{\bf Fine-tuning survivability --} whether the backdoor remains effective if the model is fine-tuned. A pre-trained model is often composed with a classifier and fine-tuned using the data from the downstream task. It is desirable to ensure that the backdoor remains effective after fine-tuning.

-- Latent Backdoor (\lb)\mcite{latent-backdoor} accounts for the impact of downstream fine-tuning by optimizing $g$ with respect to latent representations rather than final predictions. Specifically, it instantiates \meq{eq:opt} with the following loss function:
$\ell(g(x \oplus r), t) = \Delta(g(x \oplus r), \phi_t)$,
where $\Delta$ measures the difference of two latent representations and $\phi_t$ denotes the average representation of class $t$, defined as $\phi_t = \arg\min_{\phi} \sE_{(x, t) \in \mathcal{T}} [g(x)] $.

\vspace{2pt}
{\bf Defense adaptivity --}  whether the attack is optimizable to evade possible defenses. For the attack to be effective, it is essential to optimize the evasiveness of the trojan model and the trigger input with respect to the deployed defenses.

-- Adversarial-Backdoor-Embedding (\abe)\mcite{adv-backdoor} accounts for possible defenses in forging $\sg$. In solving \meq{eq:opt}, \abe also optimizes the indistinguishability of the latent representations of trigger and clean inputs. Specifically, it uses a discriminative network $d$ to predict the representation of a given input $x$ as trigger or clean. Formally, the loss is defined as $\Delta(d\circ g(x), b(x))$, where $b(x)$ encodes whether $x$ is trigger or clean, while $\sg$ and $d$ are trained using an adversarial learning framework\mcite{Goodfellow:2014:nips}.

\vspace{2pt}
{\bf Multi-optimization --} whether the attack is optimizable with respect to multiple objectives listed above. 

-- Input-Model Co-optimization (\imc)\mcite{imc} is motivated by the mutual-reinforcement effect between $r$ and $\uf$: optimizing one amplifies the effectiveness of the other. Instead of solving \meq{eq:opt} by first pre-defining $r$ and then optimizing $\uf$, \imc optimizes $r$ and $\uf$ jointly, which enlarges the search spaces for $r$ and $\uf$, leading to attacks satisfying multiple desiderata (\meg,\,fine-tuning survivability and defense adaptivity).

\begin{table*}[!ht]{\footnotesize
    \centering
    \renewcommand{\arraystretch}{1.25}
    \setlength{\tabcolsep}{2pt}
    \begin{tabular}{r|c|c|c|c|c|c|l}
    \multirow{2}{*}{\bf Neural Backdoor Defense} & \multirow{2}{*}{\bf Category} & \multicolumn{2}{c|}{\bf Mitigation} & \multicolumn{3}{c|}{\bf Detection Target} & \multirow{2}{*}{\bf Design Rationale} \\
    \cline{3-7}
    & & {\bf Input} & {\bf Model} & {\bf Input} & {\bf Model} & {\bf Trigger} & \\
    \hline
    \hline
    Randomized-Smoothing (\rands)\mcite{randomized-smoothing} & \multirow{3}{*}{\minitab[c]{Input\\Reformation}} & $\checkmark$ & & & & & $\gA$'s fidelity ($x$'s and $\ux$'s surrounding class boundaries) \\
    Down-Upsampling (\du)\mcite{feature-squeeze} & & $\checkmark$ & & & & & $\gA$'s fidelity ($x$'s and $\ux$'s high-level features) \\
    Manifold-Projection (\mmp)\mcite{magnet} & & $\checkmark$ & & & & & $\gA$'s fidelity ($x$'s and $\ux$'s manifold projections) \\
    \hline
    Activation-Clustering (\ac)\mcite{active-clustering} & \multirow{4}{*}{\minitab[c]{Input\\Filtering}} & & & $\checkmark$ & & & distinct activation patterns of $\{x\}$ and $\{\ux\}$ \\
    Spectral-Signature (\mss)\mcite{spectral-signature} & & & & $\checkmark$ & & & distinct activation patterns of $\{x\}$ and $\{\ux\}$ (spectral space) \\
    STRIP (\strip)\mcite{strip} & & & & $\checkmark$ & & & distinct self-entropy of $x$'s and $\ux$'s mixtures with clean inputs \\
    NEO (\neo)\mcite{neo} & & & & $\checkmark$ & & & sensitivity of $\uf$'s prediction to trigger perturbation \\
    \hline
    Adversarial-Retraining (\at)\mcite{madry:iclr:2018} & Model & & $\checkmark$ & & & & $\gA$'s fidelity ($x$'s and $\ux$'s surrounding class boundaries) \\
    Fine-Pruning (\fp)\mcite{fine-pruning} & Sanitization & & $\checkmark$ & & & & $\gA$'s use of neurons rarely activated by clean inputs \\
    \hline
    NeuralCleanse (\nc)\mcite{neural-cleanse} & \multirow{6}{*}{\minitab[c]{Model\\Inpsection}} & & & & $\checkmark$ & $\checkmark$ & abnormally small perturbation from other classes to $\ay$ in $\af$\\
    DeepInspect (\di)\mcite{deep-inspect} & & & & & $\checkmark$ & $\checkmark$ & abnormally small perturbation from other classes to $\ay$ in $\uf$\\
    TABOR (\tabor)\mcite{tabor} & & & & & $\checkmark$ & $\checkmark$ & abnormally small perturbation from other classes to $\ay$ in $\af$ \\
    NeuronInspect (\ninspect)\mcite{neuron-inspect} & & & & & $\checkmark$ & & distinct explanations of $f$ and $\uf$ with respect to clean inputs \\
    ABS (\abs)\mcite{abs} & & & & &$\checkmark$ &$\checkmark$ & $\gA$'s use of neurons elevating $\ay$'s prediction \\
    \end{tabular}
    \caption{Summary of representative neural backdoor defenses currently implemented in \system ($\gA$ -- backdoor attack, $x$ -- clean input, $x^*$ -- trigger input, $f$ -- benign model, $f^*$ -- trojan model, $\ay$ -- target class) \label{tab:defense_summary}}}
    \end{table*}

\subsection{Attack Performance Metrics}
\label{sec:attack-metric}

Currently, \system incorporates 6 metrics to assess the effectiveness, evasiveness, and transferability of given attacks.

\vspace{1pt}
{\underline{Attack success rate}} (\asr) -- which measures the likelihood that trigger inputs are classified to the target class $\ay$:
\begin{align}
\text{\small Attack Success Rate (\asr)} = \frac{\textrm{\small \# successful trials}}{\textrm{\small \# total trials}}
\end{align}
Typically, higher \asr indicates more effective attacks.

\vspace{1pt}
{\underline{Trojan misclassification confidence}} (\tmc) -- which is the average confidence score assigned to class $\ay$ of trigger inputs in successful attacks. Intuitively, \tmc complements \asr and measures attack efficacy from another perspective. For two attacks with the same \asr, we consider the one with higher \tmc a stronger one.

\vspace{1pt}
{\underline{Clean accuracy drop}} (\cad) -- which measures the difference of the classification accuracy of benign and trojan models; \cad measures whether the attack directs its influence to trigger inputs only.

\vspace{1pt}
{\underline{Clean classification confidence}} (\ccc) -- which is the average confidence assigned to the ground-truth classes of clean inputs; \ccc complements \cad by measuring attack specificity from the perspective of classification confidence.

\vspace{1pt}
{\underline{Efficacy-specificity AUC}} (\auc) -- which quantifies the aggregated trade-off between attack efficacy (measured by \asr) and attack specificity (measured by \cad). As revealed in\mcite{imc}, there exists an intricate balance: at a proper cost of specificity, it is possible to significantly improve efficacy, and vice versa; \auc measures the area under the \asr-\cad curve. Intuitively, smaller \auc implies a more significant trade-off effect.

\vspace{1pt}
{\underline{Neuron-separation ratio}} (\nsr) -- which measures the intersection between neurons activated by clean and trigger inputs. In the penultimate layer of the model, we find $\gN_\mathrm{c}$ and $\gN_\mathrm{t}$, the top-$k$ active neurons with respect to clean and trigger inputs, respectively, and calculate their jaccard index:
\begin{align}
\text{\small Neuron Separation Ratio (\nsr)} = 1- \frac{|\gN_\mathrm{t} \cap \gN_\mathrm{c}|}{|\gN_\mathrm{t} \cup \gN_\mathrm{c}|}
\end{align}
Intuitively, \nsr compares the neural activation patterns of clean and trigger inputs.

\subsection{Defense Library}

The existing defenses against neural backdoors, according to their strategies, can be classified into 4 major categories, as summarized in Table\mref{tab:defense_summary}. Notably, we focus on the setting of transfer learning or outsourced training, which precludes certain other defenses such as purging poisoning training data\mcite{certified-purging}. Next, we detail the 14 representative defenses currently implemented in \system.

\vspace{2pt}
{\bf Input reformation --} which, before feeding an incoming input to the model, first reforms it to mitigate the influence of the potential trigger, yet without explicitly detecting whether it is a trigger input.
It typically exploits the high fidelity of attack $\gA$, that is, $\gA$ tends to retain the perceptual similarity of a clean input $x$ and its trigger counterpart $\ux$.

-- Randomized-Smoothing (\rands)\mcite{randomized-smoothing} exploits the premise that $\gA$ retains the similarity of $x$ and $\ux$ in terms of their surrounding class boundaries and classifies an input by averaging the predictions within its vicinity (via adding Gaussian noise).

-- Down-Upsampling (\du)\mcite{feature-squeeze} exploits the premise that $\gA$ retains the similarity of $x$ and $\ux$ in terms of their high-level features while the trigger $r$ is typically not perturbation-tolerant. By downsampling and then upsampling $\ux$, it is possible to mitigate $r$'s influence.

-- Manifold-Projection (\mmp)\mcite{magnet} exploits the premise that $\gA$ retains the similarity of $x$ and $\ux$ in terms of their projections to the data manifold. Thus, it trains an autoencoder to learn an approximate manifold, which projects $\ux$ to the manifold.

\vspace{2pt}
{\bf Input filtering --} which detects whether an incoming input is embedded with a trigger and possibly recovers the clean input. It typically distinguishes clean and trigger inputs using their distinct characteristics.

-- Activation-Clustering (\ac)\mcite{active-clustering} distinguishes clean and trigger inputs by clustering their latent representations. While \ac is also applicable for purging poisoning data, we consider its use as an input filtering method at inference time. 

-- Spectral-Signature (\mss)\mcite{spectral-signature} exploits the similar property in the spectral space.

-- {\strip}\mcite{strip} mixes a given input with a clean input and measures the self-entropy of its prediction. If the input is trigger-embedded, the mixture remains dominated by the trigger and tends to be misclassified, resulting in low self-entropy.

-- {\neo}\mcite{neo} detects a trigger input by searching for a position, if replaced by a ``blocker'', changes its prediction, and uses this substitution to recover its original prediction.

\vspace{2pt}
{\bf Model sanitization --} which, before using a pre-trained model $f$, sanitizes it to mitigate the potential backdoor, yet without explicitly detecting whether $f$ is tampered.

-- Adversarial-Retraining (\at)\mcite{madry:iclr:2018} treats trigger inputs as one type of adversarial inputs and applies adversarial training over the pre-trained model to improves its robustness to backdoor attacks. 

-- Fine-Pruning (\fp)\mcite{fine-pruning} uses the property that the attack exploits spare model capacity. It thus prunes rarely used neurons and then applies fine-tuning to defend against pruning-aware attacks.

\vspace{2pt}
{\bf Model inspection --} which determines whether $f$ is a trojan model and, if so, recovers the target class and the potential trigger, at the model checking stage.

-- NeuralCleanse (\nc)\mcite{neural-cleanse} searches for potential triggers in each class $t$. If $t$ is trigger-embedded, the minimum perturbation required to change the predictions of the inputs in other classes to $t$ is abnormally small. 

-- DeepInspect (\di)\mcite{deep-inspect} follows a similar pipeline but uses a generative network to generate trigger candidates. 

-- {\tabor}\mcite{tabor} extends \nc by adding a new regularizer to control the trigger search space.

-- NeuronInspect (\ninspect)\mcite{neuron-inspect} exploits the property that the explanation heatmaps of benign and trojan models manifest distinct characteristics. Using the features extracted from such heatmaps, \ninspect detects trojan models as outliers.

-- {\abs}\mcite{abs} inspects $f$ to sift out abnormal neurons with large elevation difference (\mie,\,active only with respect to one specific class) and identifies triggers by maximizing abnormal neuron activation while preserving normal neuron behaviors.

\subsection{Defense Utility Metrics}
\label{sec:defense-metric}

Currently, \system incorporates 10 metrics to evaluate the robustness, utility-preservation, and genericity of given defenses. The metrics are tailored to the objectives of each defense category (\meg,\,trigger input detection).
For ease of exposition, below we consider the performance of a given defense $\gD$ with respect to a given attack $\gA$.

\vspace{1pt}
{\underline{Attack rate deduction}} (\ard) -- which measures the difference of $\gA$'s \asr before and after $\gD$. Intuitively, \ard indicates $\gD$'s impact on $\gA$'s efficacy. Intuitively, larger \ard indicates more effective defense. We also use $\gA$'s \tmc to measure $\gD$'s influence on the classification confidence of trigger inputs.

\vspace{1pt}
{\underline{Clean accuracy drop}} (\cad) -- which measures the difference of the \acc of clean inputs before and after $\gD$ is applied. It measures $\gD$'s impact on clean inputs. Note that \cad here is defined differently from its counterpart in attack performance metrics. We also use \ccc to measure $\gD$'s influence on the classification confidence of clean inputs.

\vspace{1pt}
{\underline{True positive rate}} (\tpr) -- which, for input-filtering methods, measures the performance of detecting trigger inputs.
\begin{align}
\text{\small True Positive Rate (\tpr)} = \frac{\textrm{\small \# detected trigger inputs}}{\textrm{\small \# total trigger inputs}}
\end{align}
Correspondingly, we use false positive rate (\fpr) to measure the error of misclassifying clean inputs as trigger inputs.

\vspace{1pt}
{\underline{Anomaly index value}} (\anidx) -- which measures the anomaly of trojan models in model-inspection defenses. Most existing methods (\meg,\,\mcite{neural-cleanse,deep-inspect,tabor,abs}) formalize finding trojan models as outlier detection: each class $t$ is associated with a score (\meg,\,minimum perturbation); if its score significantly deviates from others, $t$ is considered to contain a backdoor. \anidx, the absolute deviations from median normalized by median absolute deviation (\mad), provide a reliable measure for such dispersion. Typically, $t$ with \anidx larger than 2 has over 95\% probability of being anomaly.

\vspace{1pt}
{\underline{Mask L$_1$ norm}} (\mln) -- which measures the $\ell_1$-norm of the triggers recovered by model-inspection methods.

\vspace{1pt}
{\underline{Mask jaccard similarity}} (\mjs) -- which further measures the intersection between the recovered trigger and the ground-truth trigger (injected by the adversary). Let $\ssup{m}{o}$ and $\ssup{m}{r}$ be the masks of original and recovered triggers. We define \mjs as the Jaccard similarity of $\ssup{m}{o}$ and $\ssup{m}{r}$ :
\begin{align}
\text{\small Mask Jaccard Similarity (\mjs)} = \frac{| O(\ssup{m}{o}) \cap O(\ssup{m}{r})|}{|O(\ssup{m}{o}) \cup O(\ssup{m}{r})|}
\end{align}
where $O(m)$ denotes the set of non-zero elements in $m$.

\vspace{1pt}
{\underline{Average running time}} (\art) -- which measures $\gD$'s overhead. For model sanitization or inspection, which is performed offline, \art is measured as the running time per model; while for input filtering or reformation, which is executed online, \art is measured as the execution time per input.

\section{Assessment}
\label{sec:evaluation}

Leveraging \system, we conduct a systematic assessment of the existing attacks and defenses and unveil their complex design spectrum: both attacks and defenses tend to manifest intricate trade-offs among multiple desiderata. We begin by describing the setting of the evaluation.

\subsection{Experimental Setting}

\begin{table}[!ht]{\footnotesize 
\centering
\renewcommand{\arraystretch}{1.2}
\begin{tabular}{r|c|c|c|c}
{Dataset} & {\#\,Class} & {\#\,Dimension} & {Model} & {ACC} \\
\hline
\hline
\multirow{3}{*}{CIFAR10} & \multirow{3}{*}{10} & \multirow{3}{*}{32$\times$32}  & ResNet18 & 95.37\% \\
\cline{4-5}
& & & DenseNet121 & 93.84\% \\
\cline{4-5}
& & &  VGG13 & 92.44\% \\
\hline
CIFAR100 & 100 &  32$\times$32   & \multirow{4}{*}{ResNet18} & 73.97\% \\
\cline{1-3}
\cline{5-5}
GTSRB & 43 & 32$\times$32  &  & 98.18\% \\
\cline{1-3}
\cline{5-5}
ImageNet & 10 & 224$\times$224  & & 92.40\% \\
\cline{1-3}
\cline{5-5}
VGGFace2 & 20 & 224$\times$224  &  & 90.77\% \\
\end{tabular}
\caption{\acc of benign models over different datasets. \label{tab:dataset}}}
\end{table}

{\underline{Datasets} --} In the evaluation, we primarily use 5 datasets: {\cifar}\mcite{cifar}, {\ncifar}\mcite{cifar}, {\imgnet}\mcite{imagenet},  {\gtsrb}\mcite{gtsrb}, and {\vggface}\mcite{vggface2}, with their statistics summarized in Table\mref{tab:dataset}. 

\vspace{1pt}
{\underline{Models} --} We consider 3 representative DNN models: VGG\mcite{vgg}, ResNet\mcite{resnet}, and DenseNet\mcite{densenet}. Using models of distinct architectures (\meg, residual blocks versus skip connections), we factor out the influence of individual model characteristics. By default, we assume the downstream classifier comprising one fully-connected layer with softmax activation (1{\sc Fcn}). We also consider other types of classifiers, including Bayes, SVM, and Random Forest. The \acc of benign models is summarized in Table\mref{tab:dataset}.

\vspace{1pt}
{\underline{Attacks, Defenses, and Metrics} --} In the evaluation, we exemplify with 8 attacks in Table\mref{tab:attack_summary} and 12 defenses in Table\mref{tab:defense_summary}, and measure them using all the metrics in \msec{sec:attack-metric} and \msec{sec:defense-metric}. In all the experiments, we generate 10 trojan models for a given attack under each setting and 100 pairs of clean-trigger inputs with respect to each trojan model. The reported results are averaged over these cases. 

\vspace{1pt}
{\underline{Implementation} --} All the models, algorithms, and measurements are implemented in PyTorch. The default parameter setting is summarized in Table\mref{tab:attack_param} and\mref{tab:defense_param} (\msec{sec:details}).

\subsection{Attack Evaluation}
\label{sec:attack-eval}

We evaluate the existing attacks under the vanilla setting (without defenses), aiming to understand the impact of various design choices on the attack performance. Due to space limitations, we mainly report the results on \cifar and defer the results on other datasets to \msec{sec:additional}. Overall, different attacks manifest intricate trade-offs among {\em effectiveness}, {\em evasiveness}, and {\em transferability}, as detailed below.

\subsubsection{Effectiveness vs. Evasiveness (Trigger) }

We start with the effectiveness-evasiveness trade-off. Intuitively, the effectiveness measures whether the trigger inputs are successfully misclassified into the target class, while the evasiveness measures whether the trigger inputs and trojan models are distinguishable from their normal counterparts. Here, we first consider the evasiveness of triggers.

\begin{figure}[!ht]
    \centering
    \includegraphics[width=82mm]{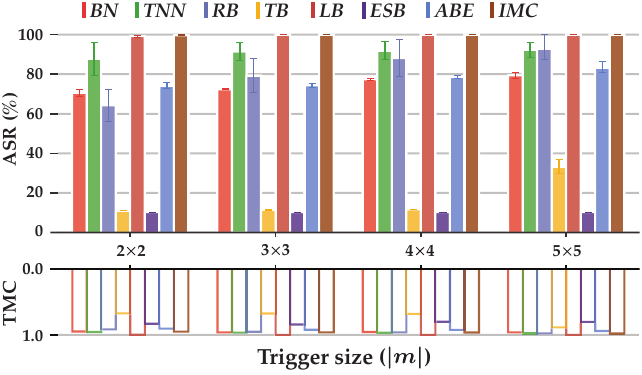}
    \caption{\asr and \tmc with respect to trigger size ($\alpha$\,=\,0.8). \label{fig:size_cifar10}}
\end{figure}

\vspace{1pt}
{\bf Trigger size --} Recall that the trigger definition comprises mask\,$m$, transparency\,$\alpha$, and pattern\,$p$. We measure how the attack effectiveness varies with the trigger size\,$|m|$. To make fair comparison, we bound the clean accuracy drop (\cad) of all the attacks below 3\% via controlling the number of optimization iterations\,$n_{\mathrm{iter}}$. Figure\mref{fig:size_cifar10} plots the attack success rate (\asr) and trojan misclassification confidence (\tmc) of various attacks under varying $|m|$ (with fixed $\alpha = 0.8$). 

Observe that most attacks seem insensitive to $|m|$: as $|m|$ varies from 2$\times$2 to 5$\times$5, the \asr of most attacks increases by less than 10\%, except \rfb with over 30\% growth. This may be attributed to its additional constraints: \rfb defines the trigger to be the reflection of another image; thus, increasing $|m|$ may improve its perturbation spaces. Compared with other attacks, \tb and \esb perform poorly because \tb aims to force inputs with random triggers to be misclassified while \esb is unable to account for trigger transparency during training. 
Also observe that the \tmc of most attacks remains close to 1.0 regardless of $|m|$.

\begin{figure}[!ht]
    \centering
    \includegraphics[width=82mm]{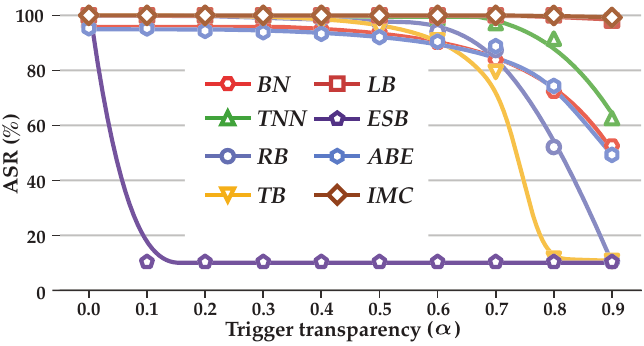}
    \caption{\asr with respect to trigger transparency ($|m|$ = 3$\times$3). \label{fig:alpha_cifar10}}
\end{figure}

\vspace{2pt}
{\bf Trigger transparency --}  Under the same setting, we evaluate the impact of trigger transparency\,$\alpha$. Figure\mref{fig:alpha_cifar10} plots the \asr of various attacks as a function of $\alpha$ ($|m|$ = 3$\times$3). 

Compared with trigger size, $\alpha$ has a more profound impact. The \asr of most attacks drops sharply once $\alpha$ exceeds 0.6, among which \tb approaches 10\% if $\alpha \geq 0.8$, and \esb works only if $\alpha$ is close to 0, due to its reliance on recognizing the trigger precisely to overwrite the model prediction. Meanwhile, \lb and \imc seem insensitive to $\alpha$. This may be attributed to that \lb optimizes trojan models with respect to latent representations (rather than final predictions), while \imc optimizes trigger patterns and trojan models jointly. Both strategies may mitigate $\alpha$'s impact. 

\begin{table}[!ht]{\footnotesize
    \centering
    \renewcommand{\arraystretch}{1.1}
\setlength{\tabcolsep}{1pt}
    \begin{tabular}{c|c|c|c|c}
        \multirow{2}{*}{Attack} & {CIFAR10}  & {CIFAR100}    & \multicolumn{2}{c}{ImageNet}  \\
        \cline{2-5}
        & $|m|$\,=\,3, $\alpha$\,=\,0.8 & $|m|$\,=\,3, $\alpha$\,=\,0.8 & $|m|$\,=\,3, $\alpha$\,=\,0 & $|m|$\,=\,7, $\alpha$\,=\,0.8 \\
        \hline
        \hline 
        \bn     & 72.4 (0.96) & 64.5 (0.96) & 90.0 (0.98) &     11.4 (0.56)       \\
        \tnn    & 91.5 (0.97) & 89.8 (0.98) & 95.2 (0.99) &   11.6 (0.62)         \\
        \rfb     & 52.1 (1.0) & 42.8 (0.95) & 94.6 (0.98) &    11.2 (0.59)        \\
        \tb     & 11.5 (0.66) & 23.4 (0.75) & 82.8 (0.97) &    11.4 (0.58)        \\
        \lb     & \cellcolor{Red}100.0 (1.0) & 97.8 (0.99) & 97.4 (0.99) & 11.4 (0.59) \\
        \esb    & 10.3 (0.43) & 1.0 (0.72) & \cellcolor{Red}100.0 (0.50) &    N/A         \\
        \abe    & 74.3 (0.91) & 67.9 (0.96) & 82.6 (0.97) &    12.00 (0.50)           \\
        \imc    & \cellcolor{Red}100.0 (1.0) & \cellcolor{Red}98.8 (0.99) & 98.4 (1.0) &  \cellcolor{Red}96.6 (0.99)       \\
    \end{tabular}
    \caption{Impact of data complexity on \asr and \tmc. \label{tab:data-comprison}}}
\end{table}

\vspace{2pt}
{\bf Data complexity --} The trade-off between attack effectiveness and trigger evasiveness is especially evident for complex data. We compare the \asr and \tmc of given attacks on different datasets, with results in Table\mref{tab:data-comprison} (more in Table\mref{tab:data-comprison2}). 

We observe that the class-space size (the number of classes) negatively affects the attack effectiveness. For example, the \asr of \bn drops by 7.9\% from \cifar to \ncifar. Intuitively, it is more difficult to force trigger inputs from all the classes to be misclassified in larger output space. Moreover, it tends to require more significant triggers to achieve comparable attack performance on more complex data. For instance, for \imc to attain similar \asr on \cifar and  \imgnet, it needs to either increase trigger size (from 3$\times$3 to 7$\times$7) or reduce trigger transparency (from 0.8 to 0.0).
\begin{mtbox}{\small Remark\,1}
    {\em \small There exists a trade-off between attack effectiveness and trigger evasiveness (in terms of transparency), which is especially evident for complex data.}
\end{mtbox}

\subsubsection{Effectiveness vs. Evasiveness (Model) }

Further, we consider the evasiveness of trojan models, which is measured by their difference from benign models in terms of classifying clean inputs. One intriguing property of the attacks is the trade-off between maximizing the attack effectiveness with respect to trigger inputs and minimizing the influence over clean inputs. Here, we characterize this trade-off via varying the fraction of trigger inputs in the training data. For each attack, we bound its \cad within 3\%, measure its highest and lowest \asr (which corresponds to its lowest and highest \cad respectively), and then normalize the \asr and \cad measures to $[0, 1]$.

\begin{figure}[!ht]
    \centering
    \includegraphics[width=82mm]{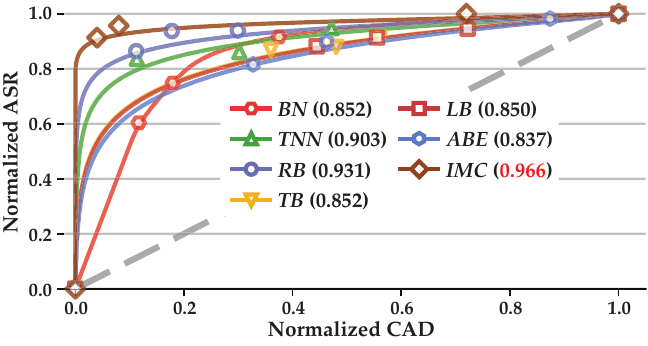}
    \caption{Trade-off between attack effectiveness and model evasiveness ($|m|\,=\,3\times3$, $\alpha$\,=\,0.8). \label{fig:auc}}
\end{figure}

Figure\mref{fig:auc} visualizes the normalized \cad-\asr trade-off. Observe that the curves of all the attacks manifest strong convexity, indicating the ``leverage'' effects\mcite{imc}: it is practical to greatly improve \asr at a disproportionally small cost of \cad. Also, observe that different attacks feature varying
Area Under the Curve (\auc). Intuitively, a smaller \auc implies a stronger leverage effect. Among all the attacks, \imc shows the smallest \auc. This may be explained by that \imc uses the trigger-model co-optimization framework, which allows the adversary to maximally optimize \asr at given \cad.

\begin{mtbox}{\small Remark\,2}
{\em \small The trade-off between attack effectiveness and model evasiveness demonstrates strong ``leverage'' effects.} 
\end{mtbox}

\subsubsection{Effectiveness vs. Transferability}

Next, we evaluate the transferability of different attacks to the downstream tasks. We consider two scenarios: \mcounter{i} the pre-training and downstream tasks share the same dataset; and \mcounter{ii} the downstream task uses a different dataset.

\vspace{2pt}
{\bf Transferability (classifier) --} In \mcounter{i}, we focus on evaluating the impact of downstream-classifier selection and fine-tuning strategy on the attacks. We consider 5 different classifiers (1/2 fully-connected layer, Bayes, SVM, and Random Forest) and 3 fine-tuning strategies (none, partial tuning, and full tuning). Notably, the adversary is unaware of such settings.

\begin{table}[!ht]
    {\footnotesize
    \centering
    \renewcommand{\arraystretch}{1.2}
  \setlength{\tabcolsep}{5pt}
    \begin{tabular}{r|c|c|c|c|c|c|c}
    \multirow{2}{*}{Attack} & \multicolumn{3}{c|}{Fine-Tuning} & \multicolumn{4}{c}{Downstream Classifier} \\
    \cline{2-8}
    & {None} & {Partial} & {Full} & {2-FCN} & {Bayes} & {SVM}  & {RF}\\
    \hline
    \hline 
    \bn & 72.4 & 72.3 & 30.4 &  72.2 &  73.5 & 64.7 & 66.0 \\
    \tnn & 91.5 &  89.6 & 27.1 &  90.8 & 90.3 & 82.9 & 81.1 \\ 
    \rfb & 79.2 & 77.0 & 12.4 &  78.3 & 76.8 & 61.5 & 63.7 \\
    \lb & {\cellcolor{Red}}100.0 & {\cellcolor{Red}}100.0 &  {\cellcolor{Red}}95.3 &  {\cellcolor{Red}}99.9 & 99.9 &  {\cellcolor{Red}}99.9 &  {\cellcolor{Red}}99.8 \\
    \imc & {\cellcolor{Red}}100.0 &  99.9 & 88.7 &   {\cellcolor{Red}}99.9 &  {\cellcolor{Red}}100.0 & {\cellcolor{Red}}99.9 &  {\cellcolor{Red}}99.8 \\ 
    \end{tabular}
    \caption{Impact of fine-tuning and downstream-model selection.\label{tab:fine-tune} }}
\end{table}

Table\mref{tab:fine-tune} compares the \asr of 5 attacks with respect to varying downstream classifiers and fine-tuning strategies. Observe that fine-tuning has a large impact on attack effectiveness. For instance, the \asr of \tnn drops by 62.5\% from partial- to full-tuning. Yet, \lb and \imc are less sensitive to fine-tuning, due to their optimization strategies. Also, note that the attack performance seems agnostic to the downstream classifier. This may be explained by that the downstream classifier in practice tends to manifest ``pseudo-linearity''\mcite{Ji:2018:ccsa} (details in \msec{sec:details}).

\vspace{2pt}
{\bf Transferability (data) --} In \mcounter{ii}, we focus on evaluating the transferability of the attacks across different datasets.

\begin{table}[!ht]{\footnotesize
\centering
\renewcommand{\arraystretch}{1.2}
\setlength{\tabcolsep}{2pt}
\begin{tabular}{c|c|c|c|c|c}

Transfer  & \multicolumn{5}{c}{Attack} \\
\cline{2-6}
Setting  & \bn  & \tnn & \rfb & \lb & \imc \\
\hline 
\hline
C $\rightarrow$ C & 94.5\,(0.99)  &100.0\,(1.0) &100.0\,(1.0) & 100.0\,(1.0) &100.0\,(1.0) \\
C $\rightarrow$ I & 8.4\,(0.29)   & 7.8\,(0.29)  & 8.6\,(0.30)   & 8.2\,(0.30)    &  \cellcolor{Red}9.4\,(0.32)    \\
\hline
I $\rightarrow$ I  & 90.0\,(0.98)  & 95.2\,(0.99) & 94.6\,(0.98) & 97.4\,(0.99)  &98.4\,(1.0)  \\
I $\rightarrow$ C &{\cellcolor{Red}}77.0\,(0.84)) & 26.9\,(0.72)         & 11.0\,(0.38)         & 10.0\,(0.38)          & 14.3\,(0.48)          \\

\end{tabular}
\caption{\asr and \tmc of transfer attacks across CIFAR10 (C) and ImageNet (I) ($|m|$\,=\,3$\times$3, $\alpha$\,=\,0.0). \label{tab:transfer}}}
\end{table}

\begin{table*}[!ht]{\footnotesize
    \centering
    \renewcommand{\arraystretch}{1.2}
    \begin{tabular}{c|c|c|c|c|c|c|c|c}
        \multirow{2}{*}{Defense} & \multicolumn{8}{c}{Attack} \\
        \cline{2-9}
             & \bn           & \tnn          & \rfb          & \tb           & \lb           & \esb          & \abe          & \imc \\
        \hline
        \hline
        -- & 93.3 (0.99) & 99.9 (1.0) & 99.8 (1.0) & 96.7 (0.99) & 100.0 (1.0) & 100.0 (0.86) & 95.3 (0.99) & 100.0 (1.0) \\
        \hline
        \hline
        \multirow{2}{*}{\rands}      & -0.5 (0.99) & -0.0 (1.0) & -0.0 -(1.0) & -0.3 (0.99) & -0.0 (1.0) & -89.1 (0.86) & -0.5 (0.99) & -0.0 (1.0)  \\
                                     & ($\pm$0.2) & ($\pm$0.0) & ($\pm$0.0) & ($\pm$0.1) & ($\pm$0.0) & ($\pm$7.3) & ($\pm$0.1) & ($\pm$0.0) \\
        \hline
        \multirow{2}{*}{\du}         & -2.2 (0.99) & -0.4 (1.0) & -5.4 (1.0) & \cellcolor{Red}-67.8 (1.0) & -4.1 (1.0) & \cellcolor{Red}-89.9 (0.86) & -0.5 (0.99) & -0.2 (1.0)      \\
                                     & ($\pm$0.7) & ($\pm$0.1) & ($\pm$1.4) & \cellcolor{Red}($\pm$12.8) & ($\pm$1.4) & \cellcolor{Red}($\pm$22.7) & ($\pm$0.3) & ($\pm$0.0)      \\
        \hline
        \multirow{2}{*}{\mmp}        & \cellcolor{Red}-6.0 (0.99) & \cellcolor{Red}-37.4 (1.0) & \cellcolor{Red}-78.6 (1.0) & -11.0 (0.99) & \cellcolor{Red} -42.6 (1.0) & -87.8 (0.86) & \cellcolor{Red}-4.6 (0.99) & \cellcolor{Red}-16.0 (1.0) \\
                                     & \cellcolor{Red}($\pm$2.1) & \cellcolor{Red}($\pm$5.5) & \cellcolor{Red}($\pm$14.2) & ($\pm$4.1) & \cellcolor{Red} ($\pm$1.5) & ($\pm$6.6) & \cellcolor{Red}($\pm$0.4) & \cellcolor{Red}($\pm$2.3) \\
        \hline
        \hline
        \multirow{2}{*}{\fp}         & -82.9 (0.60) & -86.5 (0.64) & -89.1 (0.73) & -38.0 (0.89) & -27.6 (0.82) & \cellcolor{Red}-100.0 (0.81) &-84.5 (0.64) & -26.9 (0.83) \\
                                     & ($\pm$1.8) & ($\pm$4.3) & ($\pm$2.6) & ($\pm$6.1) & ($\pm$3.7) & \cellcolor{Red}($\pm$0.0) & ($\pm$9.3) & ($\pm$4.6) \\
        \hline
        \multirow{2}{*}{\at}         & \cellcolor{Red}-83.2 (0.84) & \cellcolor{Red}-89.6 (0.85) & \cellcolor{Red}-89.8 (0.62) & \cellcolor{Red}-86.2 (0.63) & \cellcolor{Red}-90.1 (0.83) & \cellcolor{Red}-100.0 (0.86) & \cellcolor{Red}-85.3 (0.81) & \cellcolor{Red}-89.7 (0.83) \\
                                     & \cellcolor{Red}($\pm$2.2) & \cellcolor{Red}($\pm$1.9) & \cellcolor{Red}($\pm$0.7) & \cellcolor{Red}($\pm$4.5) & \cellcolor{Red}($\pm$2.8) & \cellcolor{Red}($\pm$0.0) & \cellcolor{Red}($\pm$4.4) & \cellcolor{Red}($\pm$1.8) \\
    \end{tabular}
    \caption{\ard and \tmc of attack-agnostic defenses against various attacks ($\pm$: standard deviation). \label{tab:attack-agnostic}}}
\end{table*}

We evaluate the effectiveness of transferring attacks across two datasets, \cifar and \imgnet, with results summarized in Table\mref{tab:transfer}. We have the following findings. Several attacks (\meg, \bn) are able to transfer from \imgnet to \cifar to a certain extent, but most attacks fail to transfer from \cifar to \imgnet. The finding may be justified as follows. A model pre-trained on complex data (\mie, \imgnet) tends to maintain its effectiveness of feature extraction on simple data (\mie, \cifar)\mcite{Esteva:2017:nature}; as a side effect, it may also preserve its effectiveness of propagating trigger patterns. Meanwhile, a model pre-trained on simple data may not generalize well to complex data. Moreover, compared with stronger attacks in non-transfer cases (\meg, \lb), \bn shows much higher transferability. This may be explained by that to maximize the attack efficacy, the trigger and trojan model often need to ``over-fit'' the training data, resulting in poor transferability.  

\begin{mtbox}{\small Remark\,3}
{\em \small Most attacks transfer across classifiers; however, weaker attacks demonstrate higher transferability across datasets.}
\end{mtbox}

\subsection{Defense Evaluation}
\label{sec:defense-eval}

As the defenses from different categories bear distinct objectives (\meg, detecting trigger inputs versus cleansing trojan models), below we evaluate each defense category separately.

\subsubsection{Robustness vs. Utility}

As input transformation and model sanitization mitigate backdoors in an attack-agnostic manner, while input filtering and model inspection have no direct influence on clean accuracy, we focus on evaluating attack-agnostic defenses to study the trade-off between robustness and utility preservation.

\vspace{2pt}
{\bf Robustness --} With the no-defense (vanilla) case as reference, we compare different defences in terms of attack rate deduction (\ard) and trojan misclassification confidence (\tmc), with results shown in Table\mref{tab:attack-agnostic}. We have the following observations: \mcounter{i} \mmp and \at are the most robust methods in the categories of input transformation and model sanitization, respectively. \mcounter{ii} 
\fp seems robust against most attacks except \lb and \imc, which is explained as follows: unlike attacks (\meg, \tnn) that optimize the trigger with respect to selected neurons, \lb and \imc perform optimization with respect to all the neurons,  making them immune to the pruning of \fp. 
\mcounter{iii} Most defenses are able to defend against \esb (over 85\% \ard), which is attributed to its hard-coded trigger pattern and modified DNN architecture: slight perturbation to the trigger input or trojan model may destroy the embedded backdoor. 

\begin{table}[!ht]{\footnotesize
\centering
\renewcommand{\arraystretch}{1.2}
\setlength{\tabcolsep}{0.3pt}
\begin{tabular}{c|c|c|c|c|c|c|c|c|c}
\multirow{2}{*}{\bf Defense} & \multicolumn{9}{c}{\bf Attack} \\
\cline{2-10}
& -- & \bn   & \tnn  & \rfb  & \tb   & \lb   & \esb  & \abe  & \imc \\
\hline
\hline
-- & 95.4 & 95.3 & 95.2 & 95.4 & 95.3 & 95.5 & 95.3 & 95.0 & 95.5 \\
\hline
\hline
\multirow{2}{*}{\rands}     & -0.3 & -0.6 & -0.3 & -0.4 & -0.4 & -0.3 & -0.3 & -0.4 & -0.5\\
           & ($\pm$0.2) & ($\pm$0.3) & ($\pm$0.1) & ($\pm$0.1) & ($\pm$0.3) & ($\pm$0.1) & ($\pm$0.1) & ($\pm$0.1) & ($\pm$0.2)\\
\hline
\multirow{2}{*}{\du}        & -4.0 & -4.5 & -4.5 & -4.4 & -4.3 & -4.3 & -4.0 & -4.9 & -4.6\\
           & ($\pm$0.1) & ($\pm$0.4) & ($\pm$0.3) & ($\pm$0.3) & ($\pm$0.1) & ($\pm$0.2) & ($\pm$0.2) & ($\pm$0.6) & ($\pm$0.3)\\
\hline
\multirow{2}{*}{\mmp}       & \cellcolor{Red}-11.2 & \cellcolor{Red}-11.9 &\cellcolor{Red}-11.3 & \cellcolor{Red}-10.8 & \cellcolor{Red}-11.3 & \cellcolor{Red}-11.4 & \cellcolor{Red}-11.2 & \cellcolor{Red}-11.9 & \cellcolor{Red}-11.0 \\
           & ($\pm$3.3) & ($\pm$2.1) & ($\pm$2.3) & ($\pm$1.8) & ($\pm$3.7) & ($\pm$3.2) & ($\pm$3.6) & ($\pm$3.5) & ($\pm$2.8)\\
\hline
\hline
\multirow{2}{*}{\fp}        & -0.1 & -0.2 & +0.0 & +0.0 & +0.0 &-0.2 & -0.2 & +0.3 & -0.4 \\
           & ($\pm$0.0) & ($\pm$0.0) & ($\pm$0.0) & ($\pm$0.0) & ($\pm$0.0) & ($\pm$0.1) & ($\pm$0.0) & ($\pm$0.0) & ($\pm$0.1)\\
\hline
\multirow{2}{*}{\at}        & \cellcolor{Red}-11.1 & \cellcolor{Red}-11.1 & \cellcolor{Red}-10.4 & \cellcolor{Red}-10.4 & \cellcolor{Red}-10.4 & \cellcolor{Red}-10.9 & \cellcolor{Red}-10.9 & \cellcolor{Red}-10.5 & \cellcolor{Red}-11.4 \\
           & ($\pm$4.6) & ($\pm$3.7) & ($\pm$4.4) & ($\pm$2.8) & ($\pm$3.6) & ($\pm$5.1) & ($\pm$3.0) & ($\pm$3.1) & ($\pm$3.6)\\
\end{tabular}
\caption{Impact of defenses on classification accuracy ($-$: clean model without attack/defense; $\pm$: standard deviation). \label{tab:clean-acc}}}
\end{table}

\vspace{2pt}
{\bf Utility --} We now measure the impact of defenses on the accuracy of classifying clean inputs. Table\mref{tab:clean-acc} summarizes the results. With the vanilla setting as the baseline, most defenses tend to negatively affect clean accuracy, yet with varying impact. For instance, across all the cases, \fp attains the least \cad across all the cases, mainly due to its fine-tuning; \rands and \at cause about 0.4\% and 11\% \cad, respectively. This is explained by the difference of their underlying mechanisms: although both attempt to alleviate the influence of trigger patterns, \rands smooths the prediction of an input $\bx$ over its vicinity, while \at forces the model to make consistent predictions in $x$'s vicinity. Notably, comparing with Table\mref{tab:attack-agnostic}, while \mmp and \at seem generically effective against all the attacks, they also suffer over 10\% \cad, indicating the trade-off between robustness and utility preservation.
\begin{mtbox}{\small Remark\,4}
{\em \small The design of attack-agnostic defenses faces the trade-off between robustness and utility preservation.}
\end{mtbox}

\begin{figure*}[!ht]
    \centering
    \includegraphics[width=140mm]{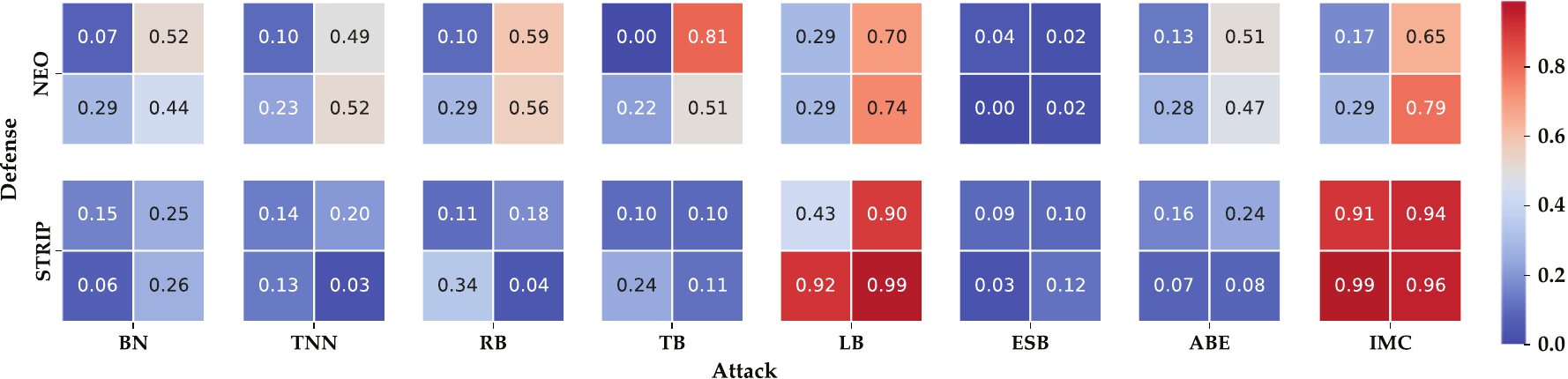}
    \caption{TPR of \neo and \strip under varying trigger definition (left:  $|m|=3\times3$, right:  $|m|=6\times6$; lower:  $\alpha=0.0$, upper:  $\alpha=0.8$). \label{fig:inference-trigger}}
\end{figure*}

\subsubsection{Detection Accuracy of Different Attacks}
\label{sec:attack-defense-eval}
We evaluate the effectiveness of input filtering by measuring its accuracy in detecting trigger inputs. 

\vspace{2pt}
{\bf Detection accuracy --} For each attack, we randomly generate 100 pairs of trigger-clean inputs and measure the true positive (\tpr) and false positive (\fpr) rates of \strip and \neo, two input filtering methods. To make comparison, we fix \fpr as 0.05 and report \tpr in Table\mref{tab:input-filtering} (statistics in \msec{sec:additional}).

\begin{table}[!ht]{\footnotesize
    \centering
    \renewcommand{\arraystretch}{1.2}
 \setlength{\tabcolsep}{0.3pt}
    \begin{tabular}{c|c|c|c|c|c|c|c|c}
        \multirow{2}{*}{Defense} & \multicolumn{8}{c}{Attack} \\
        \cline{2-9}
    & \bn & \tnn & \rfb & \tb & \lb & \esb & \abe & \imc  \\
    \hline
    \hline
   \multirow{2}{*}{\strip} & 0.07 & 0.13 & 0.34 & 0.27 & \cellcolor{Red}0.91 & 0.10 & 0.07 & \cellcolor{Red}0.99 \\
        & ($\pm$0.01) & ($\pm$0.01) & ($\pm$0.13) & ($\pm$0.08) & ($\pm$0.20) & ($\pm$0.01) & ($\pm$0.01) & ($\pm$0.02)\\
    \hline
  \multirow{2}{*}{\neo} & 0.29 & 0.23 & 0.29 & 0.36 & 0.29 & \cellcolor{Red}0.64 & 0.28 & 0.29 \\
        & ($\pm$0.09) & ($\pm$0.10) & ($\pm$0.07) & ($\pm$0.11) & ($\pm$0.06) & ($\pm$0.24) & ($\pm$0.05) & ($\pm$0.05)\\
    \end{tabular}
    \caption{TPR of \neo and \strip (FPR\,=\,0.05, $\alpha$\,=\,0.0, $\pm$\, standard deviation). \label{tab:input-filtering}}}
\end{table}

We have the following findings. \mcounter{i} \strip is particularly effective against \lb and \imc (over 0.9 \tpr). Recall that \strip detects a trigger input using the self-entropy of its mixture with a clean input. This indicates that the triggers produced by \lb and \imc effectively dominate the mixtures, which is consistent with the findings in other experiments (\mcf Figure\mref{tab:attack_summary}). \mcounter{ii} \neo is effective against most attacks to a limited extent (less than 0.3 \tpr), but especially effective against \esb (over 0.6 \tpr), due to its requirement for recognizing the trigger pattern precisely to overwrite the model prediction.  

\vspace{2pt}
{\bf Impact of trigger definition --} We also evaluate the impact of trigger definition on input filtering, with results in Figure\mref{fig:inference-trigger} (results for other defenses in \msec{sec:additional}). With fixed trigger transparency, \neo constantly attains higher \tpr under larger triggers; in comparison, \strip seems less sensitive but also less effective under larger triggers. This is attributed to the difference of their detection rationale: given input $x$, \neo searches for the ``tipping'' position in $x$ to cause prediction change, which is clearly subjective to the trigger size; while \strip measures the self-entropy of $x$'s mixture with a clean input, which does not rely on the trigger size. 
\begin{mtbox}{\small Remark\,5}
{\em \small The design of input filtering defenses needs to balance the detection accuracy with respect to different attacks.}
\end{mtbox}

\begin{table}[!ht]{\footnotesize
    \centering
    \renewcommand{\arraystretch}{1.2}
    \setlength{\tabcolsep}{0.3pt}
    \begin{tabular}{c|c|c|c|c|c|c|c|c}
            \multirow{2}{*}{Defense} & \multicolumn{8}{c}{Attack} \\
        \cline{2-9}
    & \bn & \tnn & \rfb & \tb & \lb & \esb & \abe & \imc\\
    \hline
    \hline
    \multirow{2}{*}{\nc} & 3.08 & 2.69 & 2.48 & 2.44 &2.12 & 0.04 & 2.67 & 1.66 \\
        & ($\pm$0.65) & ($\pm$0.47) & ($\pm$0.51) & ($\pm$0.38) & ($\pm$0.20) & ($\pm$0.02) & ($\pm$0.51) & ($\pm$0.25)\\
    \hline
    \multirow{2}{*}{\di} & 0.54 & 0.46 & 0.39 & 0.29 & 0.21 & 0.01 & 0.76 & 0.26 \\
        & ($\pm$0.06) & ($\pm$0.04) & ($\pm$0.04) & ($\pm$0.03) & ($\pm$0.04) & ($\pm$0.00) & ($\pm$0.10) & ($\pm$0.03)\\
    \hline
    \multirow{2}{*}{\tabor} & \cellcolor{Red}3.26 & 2.49 & 2.32 & 2.15 & 2.01 & 0.89 & 2.44 & 1.89 \\
        & ($\pm$0.77) & ($\pm$0.49) & ($\pm$0.51) & ($\pm$0.29) & ($\pm$0.63) & ($\pm$0.04) & ($\pm$0.22) & ($\pm$0.19)\\
    \hline
    \multirow{2}{*}{\ninspect} & 1.28 & 0.59 & 0.78 & 1.11 & 0.86 & 0.71 & 0.41 & 0.52 \\
        & ($\pm$0.21) & ($\pm$0.11) & ($\pm$0.06) & ($\pm$0.34) & ($\pm$0.87) & ($\pm$0.10) & ($\pm$0.05) & ($\pm$0.13)\\
    \hline
    \multirow{2}{*}{\abs} & 3.02 & \cellcolor{Red}4.16 & \cellcolor{Red}4.10 &  \cellcolor{Red}15.55 &  \cellcolor{Red}2.88 & \cellcolor{Gray} & \cellcolor{Red}8.45 & \cellcolor{Red}3.15 \\
        & ($\pm$0.81) & ($\pm$1.33) & ($\pm$1.27) & ($\pm$6.59) & ($\pm$0.25) & \cellcolor{Gray} & ($\pm$3.22) & ($\pm$0.43)\\
    \hline
    \end{tabular}
    \caption{\footnotesize \anidx of clean models and trojan models by various attacks. \label{tab:anidx}}}
\end{table}

\subsubsection{Detection Accuracy vs. Recovery Capability}

We evaluate model-inspection defenses in terms of their effectiveness of \mcounter{i} identifying trojan models and \mcounter{ii} recovering trigger patterns. 

\vspace{2pt}
{\bf Detection Accuracy --} Given defense $\gD$ and model $f$, we measure the anomaly index value (\anidx) of all the classes; if $f$ is a trojan model, we use the \anidx of the target class to quantify $\gD$'s \tpr of detecting trojan models and target classes; if $f$ is a clean model, we use the largest \anidx to quantify $\gD$'s \fpr of misclassifying clean models.

\begin{table*}[!ht]{\footnotesize
    \centering
    \renewcommand{\arraystretch}{1.2}
  \setlength{\tabcolsep}{3pt}
    \begin{tabular}{c|c c|c c|c c|c c|c c|c c|c c|c c}
        \multirow{3}{*}{Defense} & \multicolumn{16}{c}{Attack} \\
        \cline{2-17}
       & \multicolumn{2}{c|}{\bn}    & \multicolumn{2}{c|}{\tnn}     & \multicolumn{2}{c|}{\rfb}   & \multicolumn{2}{c|}{\tb}  & \multicolumn{2}{c|}{\lb}  & \multicolumn{2}{c|}{\esb}       & \multicolumn{2}{c|}{\abe}         & \multicolumn{2}{c}{\imc}   \\
       \cline{2-17}
       & MLN & MJS & MLN & MJS & MLN & MJS & MLN & MJS & MLN & MJS & MLN & MJS & MLN & MJS & MLN & MJS  \\
        \hline
        \hline
        \nc  &  \cellcolor{Blue}4.98  & 0.55   &   \cellcolor{Blue}4.65 & \cellcolor{Red}0.70      &  \cellcolor{Blue}2.64 & \cellcolor{Red}0.89   &  \cellcolor{Blue}3.53 & \cellcolor{Gray}    &  \cellcolor{Blue}7.52 & 0.21  & 35.16 & 0.00  & 5.84 & \cellcolor{Red}0.42         &  \cellcolor{Blue}8.63 & 0.13 \\
        \di  &  9.65 & 0.25   & 6.88 & 0.17     & 4.77 & 0.30  & 8.44 & \cellcolor{Gray}  & 20.17 & 0.21 &  \cellcolor{Blue}0.00 & \cellcolor{Red}0.06  & 10.21 & 0.30         & 12.78 & 0.25   \\
        \tabor & 5.63 & \cellcolor{Red}0.70   & 4.47 & 0.42     & 3.03 & 0.70  & 3.67 & \cellcolor{Gray}  & 7.65 & 0.21 & 43.37 & 0.00  &  \cellcolor{Blue}5.65 & \cellcolor{Red}0.42         & 8.69 & 0.13    \\
        \abs   &  17.74 & 0.42   &  17.91 & 0.55     &  17.60 & 0.70  &  16.00 & \cellcolor{Gray} &  17.29 & \cellcolor{Red}0.42 & \cellcolor{Gray}  &\cellcolor{Gray}   &  17.46 & 0.31  &  17.67 & \cellcolor{Red}0.31   \\
    \end{tabular}
    \caption{\mln and \mjs of triggers recovered by model-inspection defenses with respect to various attacks (Note: as the trigger position is randomly chosen in \tb, its \mjs is un-defined). \label{tab:trigger-recover} }}
\end{table*}

 The results are shown in Table\mref{tab:anidx}. We observe: \mcounter{i} compared with other defenses, \abs is highly effective in detecting trojan models (with largest \anidx), attributed to its neuron sifting strategy; \mcounter{ii} \imc seems evasive to most defenses (with \anidx below 2), explainable by its trigger-model co-optimization strategy that minimizes model distortion; \mcounter{iii} most model-inspection defenses are either ineffective or inapplicable against \esb, as it keeps the original DNN intact but adds an additional module. This contrasts the high effectiveness of other defenses against \esb (\mcf Table\mref{tab:attack-agnostic}).

\vspace{2pt}
{\bf Recovery Capability --} For successfully detected trojan models, we further evaluate the trigger recovery of various defenses by measuring the mask $\ell_1$ norm (\mln) of recovered triggers and mask jaccard similarity (\mjs) between the recovered and injected triggers, with results shown in Table\mref{tab:trigger-recover}. While the ground-truth trigger has \mln\,=\,9 ($\alpha$\,=\,0.0, $|m|$\,=\,3$\times$3), most defenses recover triggers of varying \mln and non-zero \mjs, indicating that they recover triggers different from, yet overlapping with, the injected ones. In contrast to Table\mref{tab:anidx}, \nc and \tabor outperform \abs in trigger recovery, which may be explained by that while \abs relies on the most abnormal neuron to recover the trigger, the actual trigger may be embedded into multiple neurons. This may also be corroborated by that \abs attains the highest \mjs on \lb and \imc, which tend to generate triggers embedded in a few neurons (Table\mref{tab:input-filtering}).
\begin{mtbox}{\small Remark\,6}
{\em \small The design of model-inspection defenses faces the trade-off between the accuracy of detecting trojan models and the effectiveness of recovering trigger patterns.}
\end{mtbox}

\subsubsection{Execution Time}
We compare the overhead of various defenses by measuring their \art (\msec{sec:defense-metric}) on a NVIDIA Quodro RTX6000. The results are listed in Table\mref{tab:time}. Note that online defenses (\meg, \strip) have negligible overhead, while offline methods (\meg, \abs) require longer but acceptable running time (10$^3$$\sim$10$^4$ seconds).

\begin{table}[!ht]{\footnotesize
    \centering
    \renewcommand{\arraystretch}{1.2}
    \setlength{\tabcolsep}{3pt}
    \begin{tabular}{c|c|c|c|c}
   \mmp &  \neo  & \strip &  \at &  \fp \\
   \hline
  2.4$\times$10$^1$  &   7.7$\times$10$^0$    &   1.8$\times$10$^{-1}$  &  1.7$\times$10$^4$  & 2.1$\times$10$^3$ \\
  \hline
  \hline

   \nc & \tabor & \abs & \ninspect & \di  \\
   \hline
 1.8$\times$10$^3$ & 4.2$\times$10$^3$  & 1.9$\times$10$^3$ & 4.6$\times$10$^1$ & 4.1$\times$10$^2$  
    \end{tabular}
    \caption{\footnotesize Running time of various defenses (second). \label{tab:time}}}
\end{table}

\begin{mtbox}{\small Remark\,7}
{\em \small Most defenses have marginal execution overhead with respect to practical datasets and models.}
\end{mtbox}

\subsection{Summary}
Although the defense from different categories bear distinct objectives (\meg, detecting trigger inputs versus cleansing trojan models), the evaluation above leads to the following observations: \mcounter{i}  attack-agnostic defenses often face a dilemma of trade-off between robustness and accuracy: input transformation retains high accuracy but is often ineffective against most attacks; model sanitization is 
effective to mitigate neural backdoors but at the cost of significant accuracy drop; \mcounter{ii} input-filtering is computationally efficient but only effective against a limited set of attacks; \mcounter{iii} model-inspection requires extensive optimization but the recovered trigger is able to serve as a guidance for possible backdoor unlearning. These observations may provide guidance for choosing suitable defense strategies for given application scenarios.

\section{Exploration}
\label{sec:exploration}

Next, we examine the current practices of operating backdoor attacks and defenses and explore potential improvement. 

\subsection{Attack -- Trigger}

We first explore improving the trigger definition by answering the following questions.

\vspace{2pt}
{RQ$_1$:\,{\em Is it necessary to use large triggers?}} -- It is found in \msec{sec:attack-eval} that attack efficacy seems insensitive to trigger size. We now consider the extreme case that the trigger is defined as a single pixel and evaluate the efficacy of different attacks (constrained by \cad below 5\%), with results show in Table\mref{tab:one-pixel}. Note that the trigger definition is inapplicable to \esb, due to its requirement for trigger size. 

\begin{table}[!ht]{\footnotesize
\centering
\renewcommand{\arraystretch}{1.2}
\setlength{\tabcolsep}{3pt}
\begin{tabular}{c|c|c|c|c|c|c|c}
    \bn & \tnn & \rfb & \tb & \lb & \esb & \abe & \imc \\
\hline
\hline 
95.1 & 98.1 & 77.7 & 98.0 & \cellcolor{Red}100.0 & \cellcolor{Gray} & 90.0 & 99.7 \\
(0.99)  & (0.96) & (0.96) &(0.99) & \cellcolor{Red}(0.99) & \cellcolor{Gray} & (0.97) & (0.99) 
\end{tabular}
\caption{\footnotesize \asr and \tmc of single-pixel triggers ($\alpha$\,=\,0.0, \cad\,$\leq$\,5\%).\label{tab:one-pixel} }}
\end{table}

Note that single-pixel adversarial attacks have been explored in the literature\mcite{Su:2019:tevc}; however, its study in the context of backdoor attacks is fairly limited. While it is mentioned in blind backdoor attacks\mcite{blind-backdoor}, the discussion is limited to the specific attack and does not explore the global pattern of neural backdoors. Interestingly, with single-pixel triggers, most attacks attain \asr comparable with the cases of larger triggers (\mcf Figure\mref{fig:size_cifar10}). This implies the existence of universal, single-pixel perturbation\mcite{universal-perturbation} with respect to trojan models (but not clean models!), highlighting the mutual-reinforcement effects between trigger inputs and trojan models\mcite{imc}. 

\begin{mtbox}{\small Remark\,8}
    {\em \small There often exists universal, single-pixel perturbation with respect to trojan models (but not clean models).}
\end{mtbox}

\vspace{2pt}
{RQ$_2$:\,{\em Is it necessary to use regular-shaped triggers?}} -- The triggers in the existing attacks are mostly regular-shaped (\meg,\,square), which seems a common design choice. We explore the impact of trigger shape on attack efficacy. We fix $|m|$\,=\,9 but select the positions of $|m|$ pixels independently and randomly. Table\mref{tab:scattered} compares \asr under the settings of regular and random triggers. 

\begin{table}[!ht]{\footnotesize
\centering
\renewcommand{\arraystretch}{1.2}
\begin{tabular}{c|c|c|c|c|c}
{Trigger} & \bn & \tnn & \rfb & \lb & \imc\\
\hline
\hline
Regular & 72.4 & 91.5 & 79.2 & 100.0 & 100.0 \\
Random & 97.6 & 98.5 & 92.7 & 97.6 & 94.5 \\
\end{tabular}
\caption{Comparison of regular and random triggers. \label{tab:scattered}}}
\end{table}

Except for \lb and \imc which already attain extremely high \asr under the regular-trigger setting, all the other attacks achieve higher \asr under the random-trigger setting. For instance, the \asr of \bn increases by 25.2\%. This may be explained by that lifting the spatial constraint on the trigger entails a larger optimization space for the attacks.
\begin{mtbox}{\small Remark\,9}
    {\em \small Lifting spatial constraints on trigger patterns tends to lead to more effective attacks.}
\end{mtbox}
    
\vspace{2pt}
{RQ$_3$:\,{\em Is the ``neuron-separation'' guidance effective?}} -- A common search strategy for trigger patterns is using the neuron-separation guidance: searching for triggers that activate neurons rarely used by clean inputs\mcite{trojannn}. Here, we validate this guidance by measuring the \nsr (\msec{sec:attack-metric}) of benign and trojan models before and after \fp, as shown in Table\mref{tab:jac-index}.

\begin{table}[!ht]{\footnotesize
    \centering
    \renewcommand{\arraystretch}{1.2}
\setlength{\tabcolsep}{3pt}
    \begin{tabular}{c|c|c|c|c|c|c|c}
           {Fine-Pruning}      & -- & \bn   & \tnn  & \rfb  & \lb   & \abe  & \imc  \\
        \hline
        \hline
        Before  & 0.03 & 0.59 & 0.61 & \cellcolor{Red}0.65 & 0.61 & 0.54 & 0.64 \\
        After & 0.03 & 0.20 & 0.19 & 0.27 & 0.37 & 0.18 & \cellcolor{Red}0.38 \\
    \end{tabular}
    \caption{\nsr of benign and trojan models before and after \fp. \label{tab:jac-index}}}
\end{table}

Across all the cases, compared with its benign counterpart, the trojan model tends to have higher \nsr, while fine-tuning reduces \nsr significantly. More effective attacks (\mcf Figure\mref{tab:attack_summary}) tend to have higher \nsr (\meg, \imc). We thus conclude that the neuron-separation heuristic is in general valid. 

\begin{mtbox}{\small Remark\,10}
    {\em \small The separation between the neurons activated by clean and trigger inputs is an indicator of attack effectiveness.}
\end{mtbox}

\subsection{Attack -- Optimization}

We now examine the optimization strategies used by the existing attacks and explore potential improvements.

\vspace{2pt}
{RQ$_4$:\,{\em Is it necessary to start from benign models?}} -- To forge a trojan model, a common strategy is to re-train a benign, pre-trained model. Here, we challenge this practice by evaluating whether re-training a benign model leads to more effective attacks than training a trojan model from scratch.

\begin{table}[!ht]{\footnotesize
\centering
\renewcommand{\arraystretch}{1.2}
\setlength{\tabcolsep}{3pt}
\begin{tabular}{c|c|c|c|c|c|c}
\multicolumn{2}{c|}{Training Strategy} & \bn & \tnn & \rfb & \lb & \imc \\
\hline
\hline
\multirow{2}{*}{Benign model re-training} & \asr & 72.4 & 91.5 & 79.2 & 100.0 & 100.0 \\
& \cad & -1.3 & -0.4 & -0.6 & -0.5 & -2.8 \\
\hline
\multirow{2}{*}{Training from scratch} & \asr & 76.9 & 98.9 & 81.2 & 100.0  & 100.0 \\
& \cad & -0.7 & -0.6 & -0.7 & -0.8 & -0.9 \\
\end{tabular}
\caption{\asr and \cad of trojan models by training from scratch and re-training from benign models. \label{tab:scratch}}}
\end{table}

Table\mref{tab:scratch} compares the \asr of trojan models generated using the two strategies. Except for \lb and \imc achieving similar \asr in both settings, the other attacks observe marginal improvement if they are trained from scratch. For instance, the \asr of \tnn improves by 7.4\%. One possible explanation is as follows. Let $f$ and $\uf$ represent the benign and trojan models, respectively. In the parameter space, re-training constrains the search for $\uf$ within in $f$'s vicinity, while training from scratch searches for $\uf$ in the vicinity of a randomly initialized configuration, which may lead to better starting points. 
\begin{mtbox}{\small Remark\,11}
    {\em \small Training from scratch tends to lead to more effective attacks than benign-model re-training.}
\end{mtbox}
    
\vspace{2pt}
{RQ$_5$:\, {\em Is it feasible to exploit model architectures?}} --Most attacks train trojan models in a model-agnostic manner, ignoring their unique architectures (\meg, residual block). We explore the possibility of exploiting such features. 

\begin{figure}[!ht]
    \centering
    \includegraphics[width=82mm]{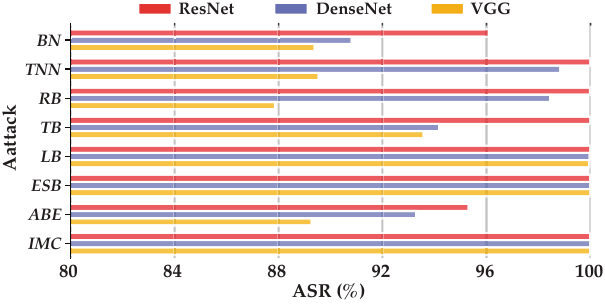}
    \caption{Impact of DNN architecture on attack efficacy. \label{fig:model_arch}}
\end{figure}

We first compare the attack performance on three DNN models, VGG, ResNet, and DenseNet, with results shown in Figure\mref{fig:model_arch}. First, different model architectures manifest varying attack vulnerabilities, ranked as ResNet\,$>$\,DenseNet\,$>$\,VGG. This may be explained as follows. Compared with traditional convolutional networks (\meg, VGG), the unique constructs of ResNet (\mie,\,residual block) and DenseNet (\mie,\,dense connection) enable more effective feature extraction, but also allow more effective propagation of trigger patterns. Second, among all the attacks, \lb, \imc, and \esb seem insensitive to model architectures, which may be attributed to the optimization strategies of \lb and \imc, and the direct modification of DNN architectures by \esb.

\begin{figure}[!ht]
    \centering
    \includegraphics[width=80mm]{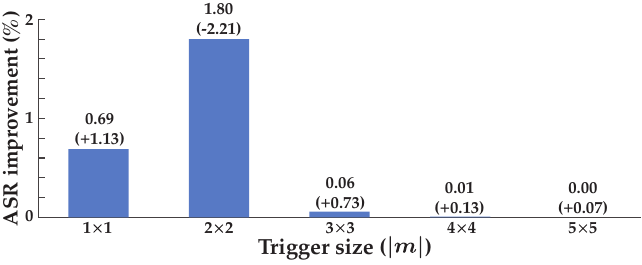}
    \caption{\asr improvement (and \cad change) by reducing skip-connection gradients ($\alpha$\,=\,0.9). \label{fig:sgm}}
\end{figure}

We then consider the skip-connect structures and attempt to improve the gradient backprop in training trojan models. In such networks, gradients propagate through both skip-connects and residual blocks. By setting the weights of gradients from skip-connects or residual blocks, it amplifies the gradient update towards inputs or model parameters\mcite{skip-connection}. Specifically, we modify the backprop procedure in \imc by setting a decay coefficient $\gamma = 0.5$ for the gradient through skip connections, with \asr improvement over normal training shown in Figure\mref{fig:sgm}.

Observe that by reducing the skip-connection gradients, it marginally improves the \asr of \imc especially for small triggers (\meg, $|m|$\,=\,2$\times$2). We consider searching for the optimal $\gamma$ to maximize attack efficacy as our ongoing work. 

\begin{mtbox}{\small Remark\,12}
    {\em \small It is feasible to exploit skip-connect structures to improve attack efficacy marginally.}
\end{mtbox}

\vspace{2pt}
{RQ$_6$:\,{\em How to mix clean and trigger inputs in training?}} -- To balance attack efficacy and specificity, the adversary often mixes clean and trigger inputs in training trojan models. There are typically three mixing strategies: \mcounter{i} dataset-level -- mixing trigger inputs $\gT_\mathrm{t}$ with clean inputs $\gT_\mathrm{c}$ directly, \mcounter{ii} batch-level -- adding trigger inputs to each batch of clean inputs during training, and \mcounter{iii} loss-level -- computing and aggregating the average losses of $\gT_\mathrm{t}$ and $\gT_\mathrm{c}$. Here, we fix the mixing coefficient $\lambda$\,=\,0.01 and compare the effectiveness of different strategies. 

\begin{table}[!ht]{\footnotesize
\centering
\renewcommand{\arraystretch}{1.2}
\begin{tabular}{c|c|c|c|c|c}
{Mixing Strategy} & \bn & \tnn & \rfb & \lb & \imc \\
\hline
\hline
Dataset-level & 59.3 & 72.2 & 46.2 & 99.6 & 92.0 \\
Batch-level & \cellcolor{Red}72.4 & \cellcolor{Red}91.5 & \cellcolor{Red}79.2 & \cellcolor{Red}100.0 & \cellcolor{Red}100.0 \\
Loss-level & 21.6 & 22.9 & 18.1 & 33.6 & 96.5  \\
\end{tabular}
\caption{Impact of mixing strategies on attack efficacy ($\alpha$\,=\,0.0, $\lambda$\,=\,0.01).\label{tab:mix}}}
\end{table}

We observe in Table\mref{tab:mix} that across all the cases, the batch-level mixing strategy leads to the highest \asr. This can be explained as follows. With dataset-level mixing, the ratio of trigger inputs in each batch tends to fluctuate significantly due to random shuffling, resulting in inferior training quality. With loss-level mixing, 
$\lambda$\,=\,0.01 results in fairly small gradients of trigger inputs, equivalent to setting an overly small learning rate. In comparison, batch-level mixing asserts every poisoning instance and its clean version must share the same batch, making the model focus more on the trigger as the classification evidence of target class.

Here, we provide a potential explanation: the loss-level mixing involves the gradient scale of poisoning data. If the loss is defined as $\mathcal{L}=\mathcal{L}_{clean}+\lambda \cdot \mathcal{L}_{poison}$ and optimization step as $\Delta=\text{lr}\cdot \frac{\partial ( \mathcal{L}_{clean}+\lambda\cdot\mathcal{L}_{poison})}{\partial \theta}$, where $\text{lr}$ is the learning rate and $\mathcal{L}_{clean}$ and $\mathcal{L}_{poison}$ are the losses on the clean and poisoning data. Observe that $\Delta=\text{lr}\cdot \frac{\partial  \mathcal{L}_{clean}}{\partial \theta} + \textbf{lr}\cdot \lambda \cdot \frac{\partial  \mathcal{L}_{poison}}{\partial \theta}$. The real gradient scale is $\textbf{lr}\cdot \lambda$ rather than $\textbf{lr}$, which makes the step size smaller than expected. 
\begin{mtbox}{\small Remark\,13}
    {\em \small Batch-level mixing tends to lead to the most effective training of trojan models.}
\end{mtbox}

\begin{figure}[!ht]
    \centering
    \includegraphics[width=75mm]{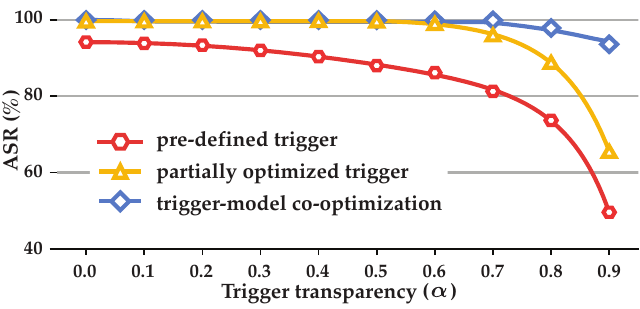}
    \caption{Impact of trigger optimization. \label{fig:optim_term}}
\end{figure}

\vspace{2pt}
{RQ$_7$:\,{\em How to optimize the trigger pattern?}} -- An attack involves optimizing both the trigger pattern and the trojan model. The existing attacks use 3 typical strategies: \mcounter{i} Pre-defined trigger -- it fixes the trigger pattern and only optimizes the trojan model. 
\mcounter{ii} Partially optimized trigger -- it optimizes the trigger pattern in a pre-processing stage and optimizes the trojan model. \mcounter{iii} Trigger-model co-optimization -- it optimizes the trigger pattern and the trojan model jointly during training. Here, we implement 3 variants of \bn that use these optimization strategies, respectively. Figure\mref{fig:optim_term} compares their \asr under varying trigger transparency. Observe that the trigger-optimization strategy has a significant impact on \asr, especially under high transparency. For instance, if $\alpha$\,=\,0.9, the co-optimization strategy improves \asr by over 60\% from the non-optimization strategy.
\begin{mtbox}{\small Remark\,14}
    {\em \small Optimizing the trigger pattern and the trojan model jointly leads to more effective attacks.}
\end{mtbox}

\subsection{Defense -- Evadability}

{RQ$_8$:\,{\em Are the existing defenses evadable?}} -- We now explore whether the existing defenses are potentially evadable by adaptive attacks. We select \imc as the basic attack, due to its flexible optimization framework, and consider \mmp, \at, \strip, and \abs as the representative defenses from the categories in Table\mref{tab:defense_summary}. Specifically, we adapt \imc to each defense.

Recall that \mmp uses an auto-encoder to downsample then upsample a given input, during which the trigger pattern tends to be blurred and loses effect. To adapt \imc to \mmp, we train a surrogate autoencoder $h$ and conduct optimization with  inputs reformed by $h$. 

Recall that \at considers trigger inputs as one type of adversarial inputs and applies adversarial training to improve model robustness against backdoor attacks. To adapt \imc to \at, during training $\uf$, we replace clean accuracy loss with adversarial accuracy loss; thus, the process is a combination of adversarial training and trojan model training, resulting in a robust but trojan model. This way, \at has a limited impact on the embedded backdoor, as the model is already robust.

Recall that \strip mixes up given inputs with clean inputs and measures the self-entropy of their predictions. Note that in the mixture, the transparency of the original trigger is doubled; yet, \strip works as the high-transparency trigger remains effective. To adapt \imc to \strip, we use trigger inputs with high-transparency triggers together with their ground-truth classes to re-train $\uf$. The re-training reduces the effectiveness of high-transparency triggers while keeping low-transparency triggers effective. 

Recall that \abs identifies triggers by maximizing abnormal activation while preserving normal neuron behavior. To adapt \imc to \abs, we integrate the cost function (Algorithm 2 in\mcite{abs}) in the loss function to train $\uf$.

\begin{figure*}[!ht]
    \centering
    \includegraphics[width=170mm]{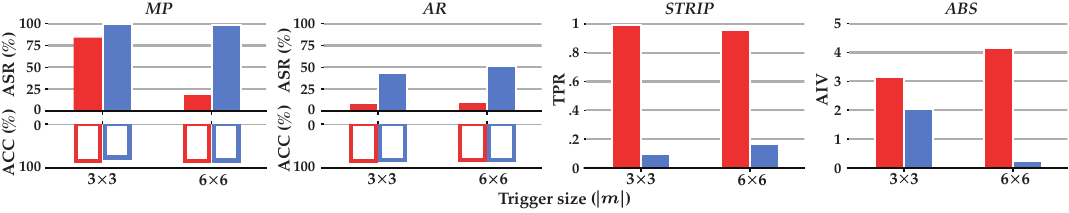}
    \caption{Performance of non-adaptive and adaptive \imc against representative defenses ($\alpha$\,=\,0.0). \label{fig:adaptation}}
\end{figure*}

We compare the efficacy of non-adaptive and adaptive \imc, as shown in Figure\mref{fig:adaptation}. Observe that across all the cases, the adaptive \imc significantly outperforms the non-adaptive one. For instance, under $|m|$\,=\,6$\times$6, it increases the \asr with respect to \mmp by 80\% and reduces the \tpr of \strip by over 0.85. Also note that a larger trigger size leads to more effective adaptive attacks, as it entails a larger optimization space.
\begin{mtbox}{\small Remark\,15}
    {\em \small Most existing defenses are potentially evadable by adaptive attacks.}
\end{mtbox}

\subsection{Defense -- Interpretability}

\begin{figure}[!ht]
    \centering
    \includegraphics[width=80mm]{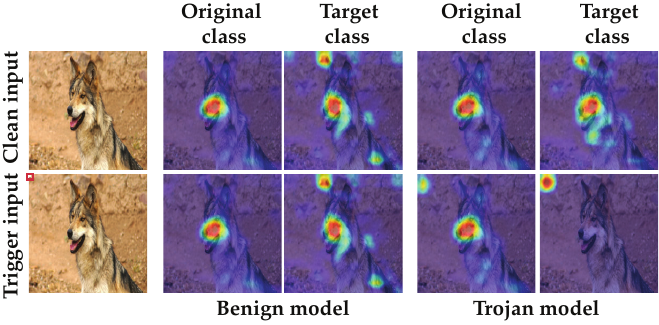}
    \caption{Sample attribution maps of clean and trigger inputs with respect to benign and trojan models ($\alpha$\,=\,0.0, \imgnet). \label{fig:example}}
\end{figure}

{RQ$_9$:\,{\em Does interpretability help mitigate backdoor attacks?}} -- The interpretability of DNNs explain how they make predictions for given inputs\mcite{selvaraju:gradcam,fong:mask}. Recent studies\mcite{Tao:nips:2018,Guo:2018:ccs} show that such interpretability helps defend against adversarial attacks. Here, we explore whether it mitigates backdoor attacks. Specifically, for a pair of benign-trojan models and 100 pairs of clean-trigger inputs, we generate the attribution map\mcite{selvaraju:gradcam} of each input with respect to both models and ground and target classes, with an example shown in Figure\mref{fig:example}. 

We measure the difference ($\ell_1$-norm normalized by image size) of attribution maps of clean and trigger inputs. Observe in Table\mref{tab:map-difference} that their attribution maps with respect to the target class differ significantly on the trojan model, indicating the possibility of using interpretability to detect the attack. This finding also corroborates recent work on using interpretability to identify possibly tampered regions in images\mcite{sentinet}. However, it may require further study whether the adversary may adapt the attack to deceive such detection\mcite{adv}.

\begin{table}[!ht]{\footnotesize
    \centering
    \renewcommand{\arraystretch}{1.2}
    \begin{tabular}{c|c|c|c}
    \multicolumn{2}{c|}{Benign model} &  \multicolumn{2}{c}{Trojan model} \\
    \hline
    {Original class} & {Target class} &  {Original class} & {Target class} \\
    \hline
    \hline
    0.08\% & 0.12\% & 0.63\% &\cellcolor{Red}8.52\% \\
    \end{tabular}
    \caption{Distance between the heatmaps of clean and trigger inputs ($\alpha$\,=\,0.0).  \label{tab:map-difference} }}
\end{table}

\begin{mtbox}{\small Remark\,16}
    {\em \small It seems promising to exploit model interpretability to enhance defense robustness.}
\end{mtbox}

\subsection{Summary}

Based on the study above, we recommend the following testing strategy for a new neural backdoor attack: \mcounter{i} attacks that optimize models only (\meg, \bn), \mcounter{ii} attacks that partially optimize triggers (\meg, \tnn), \mcounter{iii} attacks that optimize both models and triggers (\meg, \imc), and \mcounter{iv} attacks adaptive to the given defense. The increasing level of complexity gives the adversary more flexibility to optimize various settings (\meg, trigger transparency and size) to evade the defense, leading to stronger attacks.

Looking forward, the study also opens several research directions for future defenses: \mcounter{i} ensemble defenses that leverage the strengths of individual ones (\meg, input transformation and model sanitization), \mcounter{ii} defenses that involve human in the loop via interpretability, and \mcounter{iii} defenses that provide theoretical guarantees based on the invariant properties of various attacks.

\section{Limitations}
\label{sec:discussion}

First, to date \system has integrated 8 attacks and 14 defenses, representing the state of the art of neural backdoor research. Yet, as a highly active research field, a set of concurrent work has proposed new backdoor attacks/defenses\mcite{mnt,rab,benign-feature-backdoor,deepsweep,explanation-guided-poisoning-attack,tact}, which are not included in the current implementation of \system. As examples, \mcite{tact} presents a new  attack that obscures the representations of benign and trigger inputs; \mcite{explanation-guided-poisoning-attack} proposes to leverage interpretability to improve attack effectiveness; while \mcite{deepsweep} investigates data augmentation-based defenses. However, thanks to its modular design, \system can be readily extended to incorporate new attacks, defenses, and metrics. Moreover, we plan to open-source all the code and data of \system and encourage the community to contribute.

Second, to conduct a unified evaluation, we mainly consider the attack vector of re-using pre-trained trojan models. There are other attack vectors through which backdoor attacks can be launched, including poisoning victims' training data\mcite{Shafahi:2018:nips,Zhu:2019:icml} and knowledge distillation\mcite{distillation-backdoor}, which entail additional constraints for attacks or defenses. For instance, the poisoning data needs to be evasive to bypass inspection. We consider studying alternative attack vectors as our ongoing work.

Third, due to space limitations, our evaluation focuses on popular \dnn models (\meg, ResNet) and assumes fixed training/test data split. We consider evaluating the impact of model configuration and data split on neural backdoor attacks/defenses as our ongoing work.

Finally, because of the plethora of work on neural backdoors in the computer vision domain, \system focuses on the image classification task, while recent work has also explored neural backdoors in other settings, including natural language processing\mcite{schuster-humpty,acl-backdoor,lm-trojan}, reinforcement learning\mcite{trojdrl}, and federated learning\mcite{federated-backdoor,dba}. We plan to extend \system to support such settings in its future releases.

\section{Conclusion}
\label{sec:conclusion}

We design and implement \system, the first platform dedicated to assessing neural backdoor attacks/defenses in a holistic, unified, and practical manner. Leveraging \system, we conduct a systematic evaluation of existing attacks/defenses, which demystifies a number of open questions, reveals various design trade-offs, and sheds light on further improvement. We envision \system will serve as a useful benchmark to facilitate neural backdoor research.

\section*{Acknowledgment}
We thank anonymous reviewers and shepherd for valuable feedback. This work is supported by the National Science Foundation under Grant No. 1951729, 1953813, and 1953893. Any opinions, findings, and conclusions or recommendations are those of the authors and do not necessarily reflect the views of the National Science Foundation. X. Luo is partly supported by Hong Kong RGC Project (No. PolyU15222320).

\newpage
\bibliographystyle{plain}
\bibliography{bibs/aml.bib,bibs/debugging.bib,bibs/general.bib,bibs/ting.bib,bibs/optimization.bib,bibs/main.bib,bibs/interpretation.bib,bibs/ren.bib}

\newpage
\appendices

\section{Implementation Details}
\label{sec:details}

Below we elaborate on the implementation of attacks and defenses in this paper.


\subsection{Default Parameter Setting}

Table\mref{tab:attack_param} and Table\mref{tab:defense_param} summarize the default parameter setting in our empirical evaluation (\msec{sec:evaluation}).

\begin{table}[!ht]{\footnotesize 
        \centering
        \renewcommand{\arraystretch}{1.2}
        \setlength{\tabcolsep}{2pt}
        \begin{tabular}{c|l|l}
             {Attack}                    & {Parameter}                              & {Setting}             \\
            \hline
            \hline
            \multirow{5}{*}{Training} & learning rate             & 0.01                    \\
                                      & retrain epoch             & 50                      \\
                                      & optimizer                 & SGD (nesterov)          \\
                                      & momentum                  & 0.9                     \\
                                      & weight decay              & 2e-4                    \\
            \hline

            \multirow{1}{*}{\bn}      & toxic data percent        & 1\%                     \\
            \hline

            \multirow{7}{*}{\tnn}     & preprocess layer          & penultimate logits           \\ 
                                      & neuron number             & 2                       \\
                                      & preprocess optimizer      & PGD                     \\
                                      & preprocess lr             & 0.015                   \\
                                      & preprocess iter           & 20                      \\
                                      & threshold                 & 5                       \\
                                      & target value              & 10                      \\
            \hline

            \multirow{4}{*}{\rfb}     & candidate number          & 50                      \\
                                      & selection number          & 10                      \\
                                      & selection iter            & 5                       \\
                                      & inner epoch               & 5                       \\
            \hline

            \multirow{6}{*}{\lb}      & preprocess layer          & penultimate logits       \\ 
                                      & preprocess lr             & 0.1                     \\
                                      & preprocess optimizer      & Adam (tanh constrained) \\
                                      & preprocess iter           & 100                     \\
                                      & samples per class         & 1000                    \\
                                      & MSE loss weight           & 0.5                     \\
            \hline

            \multirow{6}{*}{\esb}     & TrojanNet                 & 4-layer MLP             \\
                                      & hidden neurons per layer  & 8                       \\
                                      & single layer structure    & [fc, bn, relu]          \\
                                      & TrojanNet influence       & $\alpha=$0.7            \\
                                      & amplify rate              & 100                     \\
                                      & temperature               & 0.1                     \\
            \hline

            \multirow{2}{*}{\abe}     & discriminator loss weight & $\lambda=$0.1           \\
                                      & discriminator lr          & 1e-3                    \\
            \hline

            \multirow{3}{*}{\imc}     & trigger optimizer         & PGD                     \\
                                      & PGD lr                    & $\alpha=$20/255         \\
                                      & PGD iter                  & 20                      \\
        \end{tabular}
        \caption{Attack default parameter setting. \label{tab:attack_param}}}
\end{table}

\begin{table}[!ht]{\footnotesize 
        \centering
        \renewcommand{\arraystretch}{1.2}
        \setlength{\tabcolsep}{2pt}
        \begin{tabular}{c|l|l}
            {Defense}                    & {Parameter}                              & {Setting}             \\
            \hline
            \hline

            \multirow{3}{*}{\rands}    & sample distribution                    & Gaussian            \\
                                      & sample number                          & 100                 \\
                                      & sample std                             & 0.01                \\
            \hline

            \multirow{2}{*}{\du}       & downsample filter                      & Anti Alias          \\
                                      & downsample ratio                       & 0.95                \\
            \hline

            \multirow{2}{*}{\mmp}      & training noise std                     & 0.1                 \\
                                      & structure                              & [32]                \\
            \hline

            \multirow{2}{*}{\strip}    & mixing weight                          & 0.5 (equal)         \\
                                      & sample number                          & 64                  \\
            \hline

            \multirow{3}{*}{\neo}      & sample number                          & 100                 \\
                                      & Kmeans cluster number                  & 3                   \\
                                      & threshold                              & 80                  \\
            \hline

            \multirow{5}{*}{\at}       & PGD lr                                 & $\alpha=$2/255      \\
                                      & perturbation threshold                 & $\epsilon=$8/255    \\
                                      & PGD iter                               & 7                   \\
                                      & learning rate                          & 0.01                \\
                                      & epoch                                  & 50                  \\
            \hline

            \multirow{1}{*}{\fp}       & prune ratio                            & 0.95                \\
            \hline

            \multirow{3}{*}{\nc}       & norm regularization weight             & 1e-3                \\
                                      & remask lr                              & 0.1                 \\
                                      & remask epoch per label                 & 10                  \\
            \hline

            \multirow{4}{*}{\di}       & sample dataset ratio                   & 0.1                 \\
                                      & noise dimension                        & 100                 \\
                                      & remask lr                              & 0.01                \\
                                      & remask epoch per label                 & 20                  \\
            \hline

            \multirow{6}{*}{\tabor}    & \multirow{6}{*}{regularization weight} & $\lambda_{1}=$1e-6  \\
                                      &                                        & $\lambda_{2}=$1e-5  \\
                                      &                                        & $\lambda_{3}=$1e-7  \\
                                      &                                        & $\lambda_{4}=$1e-8  \\
                                      &                                        & $\lambda_{5}=$0     \\
                                      &                                        & $\lambda_{6}=$1e-2  \\
            \hline

            \multirow{5}{*}{\ninspect} & \multirow{3}{*}{weighting coefficient} & $\lambda_{sp}=$1e-5 \\
                                      &                                        & $\lambda_{sm}=$1e-5 \\
                                      &                                        & $\lambda_{pe}=$1    \\\cline{2-3}
                                      & threshold                              & 0                   \\
                                      & sample ratio                           & 0.1                 \\
            \hline

            \multirow{8}{*}{\abs}      & sample k                               & 1                   \\
                                      & sample number                          & 5                   \\
                                      & max trojan size                        & 16                  \\
                                      & remask lr                              & 0.1                 \\
                                      & remask iter per neuron                 & 1000                \\\cline{2-3}
                                      & \multirow{3}{*}{remask weight}         & 0.1 if norm$<16$       \\
                                      &                                        & 10 if $16<$norm$<100$    \\
                                      &                                        & 100 if norm$>100$      \\
        \end{tabular}
        \caption{Defense default parameter setting. \label{tab:defense_param}}}
\end{table}

\subsection{Pseudo-linearity of downstream model}

We have shown in \msec{sec:evaluation} that most attacks seem agnostic to the downstream model. Here, we provide possible explanations. Consider a binary classification setting and a trigger input $\ax$ with ground-truth class\,``-'' and target class\,``+''.
Recall that a backdoor attack essentially shifts $\ax$ in the feature space by maximizing the quantity of
\begin{align}
\Delta_{f} = \mathbb{E}_{\mu^+}[f(\ax)] - \mathbb{E}_{\mu^-}[f(\ax)]
\end{align}
where $\mu^+$ and $\mu^-$ respectively denote the data distribution of the ground-truth positive and negative classes.

Now consider the end-to-end system $g\circ f$. The likelihood that $\ax$ is misclassified into ``+'' is given by:
\begin{align}
\Delta_{g\circ f} = \mathbb{E}_{\mu^+}[g\circ f(\ax)] - \mathbb{E}_{\mu^-}[g\circ f(\ax)]
\end{align}

One sufficient condition for the attack to succeed is that $\Delta_{g\circ f}$ is linearly correlated with $\Delta_{f}$ (\mie, $\Delta_{g\circ f} \propto \Delta_{f}$). If so, we say that the function represented by $g$ is {\em pseudo-linear}. Unfortunately, in practice, most downstream models are fairly simple (\meg, one fully-connected layer), showing pseudo-linearity. Possible reasons include: \mcounter{i} complex architectures are difficult to train especially when the training data is limited; \mcounter{ii} they imply much higher computational overhead; \mcounter{iii} the ground-truth mapping from the feature space to the output space may indeed be pseudo-linear.

\section{Additional Experiments}
\label{sec:additional}

\subsection{Attack}

\begin{figure}[!ht]
    \centering
    \includegraphics[width=82mm]{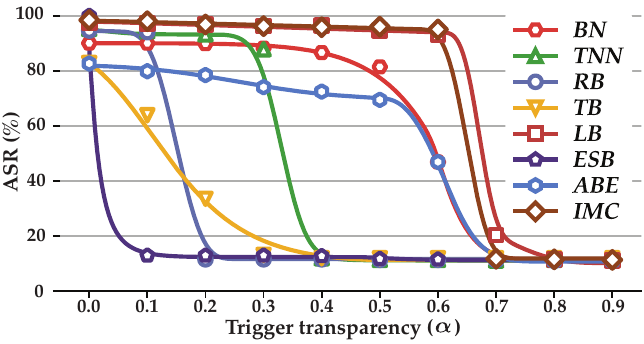}
    \caption{\asr with respect to trigger transparency ($|m|$ = 3$\times$3, ImageNet). \label{fig:alpha_imagenet}}
\end{figure}

\begin{figure}[!ht]
    \centering
    \includegraphics[width=82mm]{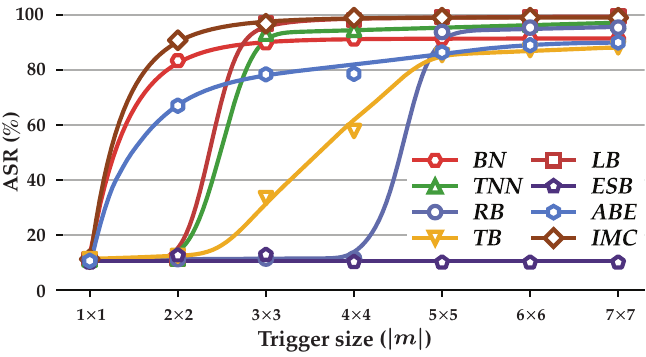}
    \caption{\asr with respect to trigger size ($\alpha$ = 0.3, ImageNet). \label{fig:size_imagenet}}
\end{figure}

Figure\mref{fig:alpha_imagenet} and \mref{fig:size_imagenet} complement the results of attack performance evaluation on ImageNet with respect to trigger size and trigger transparency in Section\mref{sec:attack-eval}. Note that Figure\mref{fig:size_imagenet} uses $\alpha=0.3$, which is more transparent than $\alpha=0.0$ used in Table\mref{tab:one-pixel}. Therefore, all attacks at $1\times 1$ trigger size are not working and their \asr are close to 10\%. This is not conflict to the observation in Table\mref{tab:one-pixel}.

The attacks tend to be sensitive to the trigger transparency but insensitive to the trigger size (claimed in Section 4.2.1). All the attacks fail under $|m|=1\times 1$ and are excluded from Figure 3 in Section 4.2.1. Table 14 and Figure 13 use different settings. Table 14: $\alpha=0.0$ on CIFAR10, Figure 13: $\alpha=0.3$ on ImageNet, which cause the difference in terms of trigger transparency and data complexity.

The attacks tend to be sensitive to the trigger transparency but insensitive to the trigger size (claimed in Section 4.2.1). $|m|=1\times 1$ is not working for all attacks and are excluded from Figure 3 in Section 4.2.1. Table 14 and Figure 13 use different settings. Table 14: $\alpha=0.0$ on CIFAR10, Figure 13: $\alpha=0.3$ on ImageNet, which cause the difference in terms of trigger transparency and data complexity.

Table\mref{tab:data-comprison2} complements the results in Table\mref{tab:data-comprison}.

\begin{table}[!ht]{\footnotesize
    \centering
    \renewcommand{\arraystretch}{1.2}
    \setlength{\tabcolsep}{2pt}
    \begin{tabular}{c|c|c|c|c|c|c|c|c}
         & \bn    & \tnn   & \rfb   & \tb    & \lb    & \esb   & \abe  & \imc   \\
         \hline
        GTSRB    & 65.63 & 71.70 & 0.94 & 0.58 & 98.42 & 68.41 & 68.41 & 97.58 \\
        CIFAR100 & 64.53  & 89.76  & 42.77 & 23.44  & 97.83  & 0.98   & 67.86 & 98.75  \\
        VGGFace2 & 85.62 & 97.30 & 92.31 & 88.75 & 98.08 & 100.00    & 72.74 & 98.43 \\
    \end{tabular}
    \caption{Impact of data complexity on \asr ($|m|=3\times 3$ and $\alpha=0.8$ for GTSRB and CIFAR100, $|m|=25\times 25$ and $\alpha=0.0$ for VGGFace2). \label{tab:data-comprison2}}}
\end{table}

\subsection{Defense}

Table\mref{tab:input-filtering2} presents more information (F1-score, precision, recall, and accuracy), which complements Table\mref{tab:input-filtering}.

\begin{table}[!ht]{\footnotesize
    \centering
    \renewcommand{\arraystretch}{1.2}
    \setlength{\tabcolsep}{3pt}
    \begin{tabular}{c|c|c|c|c|c|c|c|c|c}
  {\bf Defense}  & {\bf Measure} & \bn & \tnn & \rfb & \tb & \lb & \esb & \abe & \imc \\
    \hline
    \multirow{4}{*}{\strip} & F1 Score & 0.12 & 0.21 & 0.47 & 0.39 & \cellcolor{Red} 0.91 & 0.18 & 0.13 & \cellcolor{Red} 0.95 \\
    & Precision & 0.41 & 0.56 & 0.77 & 0.73 & 0.90 & 0.52 & 0.43 & 0.91 \\
    & Recall & 0.07 & 0.13 & 0.34 & 0.27 & 0.91 & 0.10 & 0.07 & 0.99 \\
    & Accuracy & 0.48 & 0.51 & 0.62 & 0.58 & 0.91 & 0.50 & 0.49 & 0.95 \\
    \hline
    \multirow{4}{*}{\neo} & F1 Score & 0.45 & 0.37 & 0.45 & 0.34 & 0.45 & \cellcolor{Red}0.77 & 0.43 & 0.45 \\
    & Precision & 1.00 & 1.00 & 1.00 & 0.35 & 1.00 & 0.96 & 0.90 & 1.00 \\
    & Recall & 0.29 & 0.23 & 0.29 & 0.36 & 0.29 & 0.64 & 0.28 & 0.29 \\
    & Accuracy & 0.65 & 0.62 & 0.65 & 0.36 & 0.65 & 0.81 & 0.63 & 0.65 \\
    \hline
    \end{tabular}
    \caption{Additional statistics of input filtering. \label{tab:input-filtering2}}}
\end{table}

\begin{figure}[!ht]
    \centering
    \includegraphics[width=90mm]{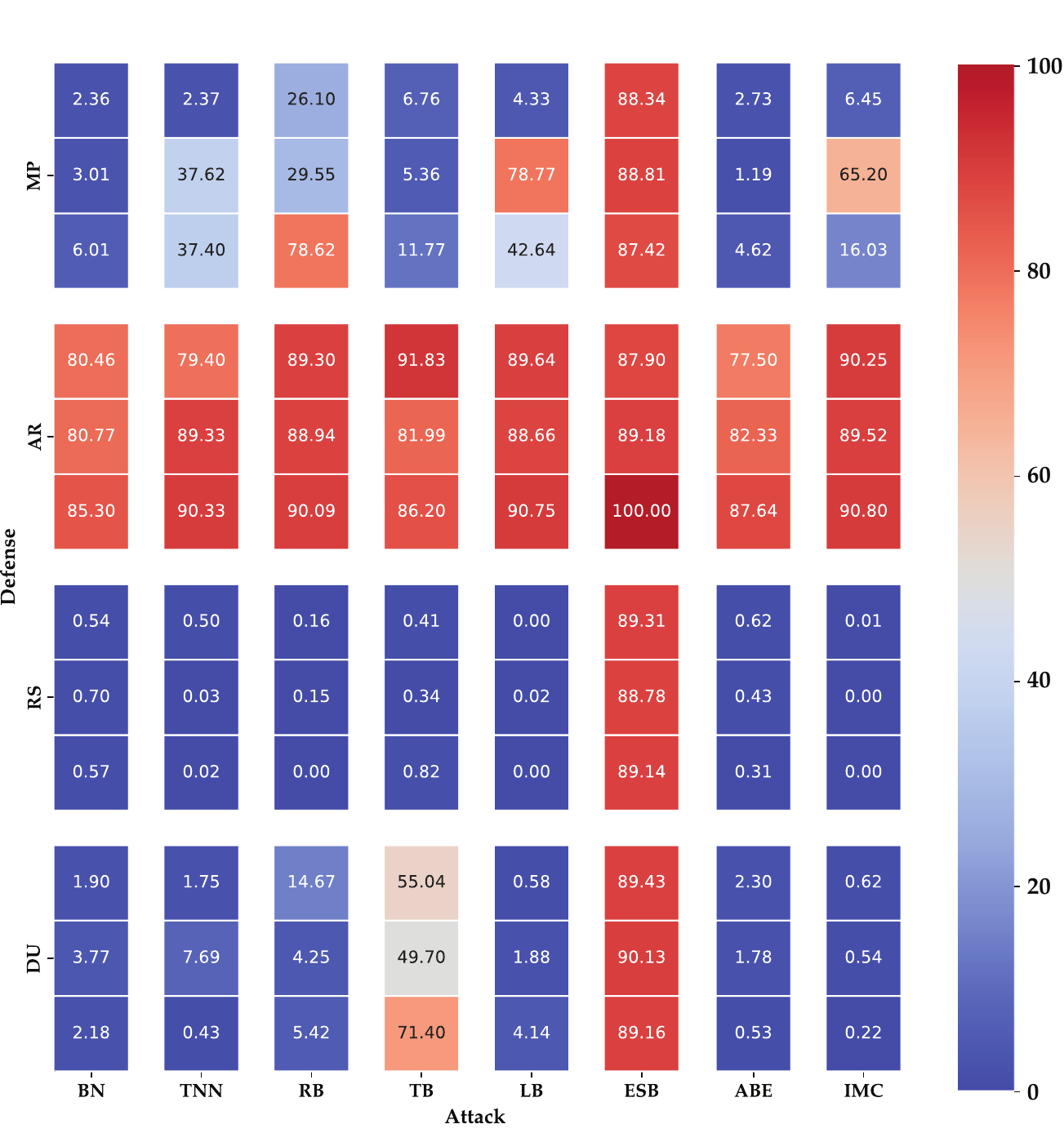}
    \caption{Impact of DNN architecture on attack-agnostic defenses (lower:  ResNet18, middle:  DenseNet121; upper:  VGG13).}
    \label{fig:defense-model-general}
\end{figure}

\begin{figure}[!ht]
    \centering
    \includegraphics[width=85mm]{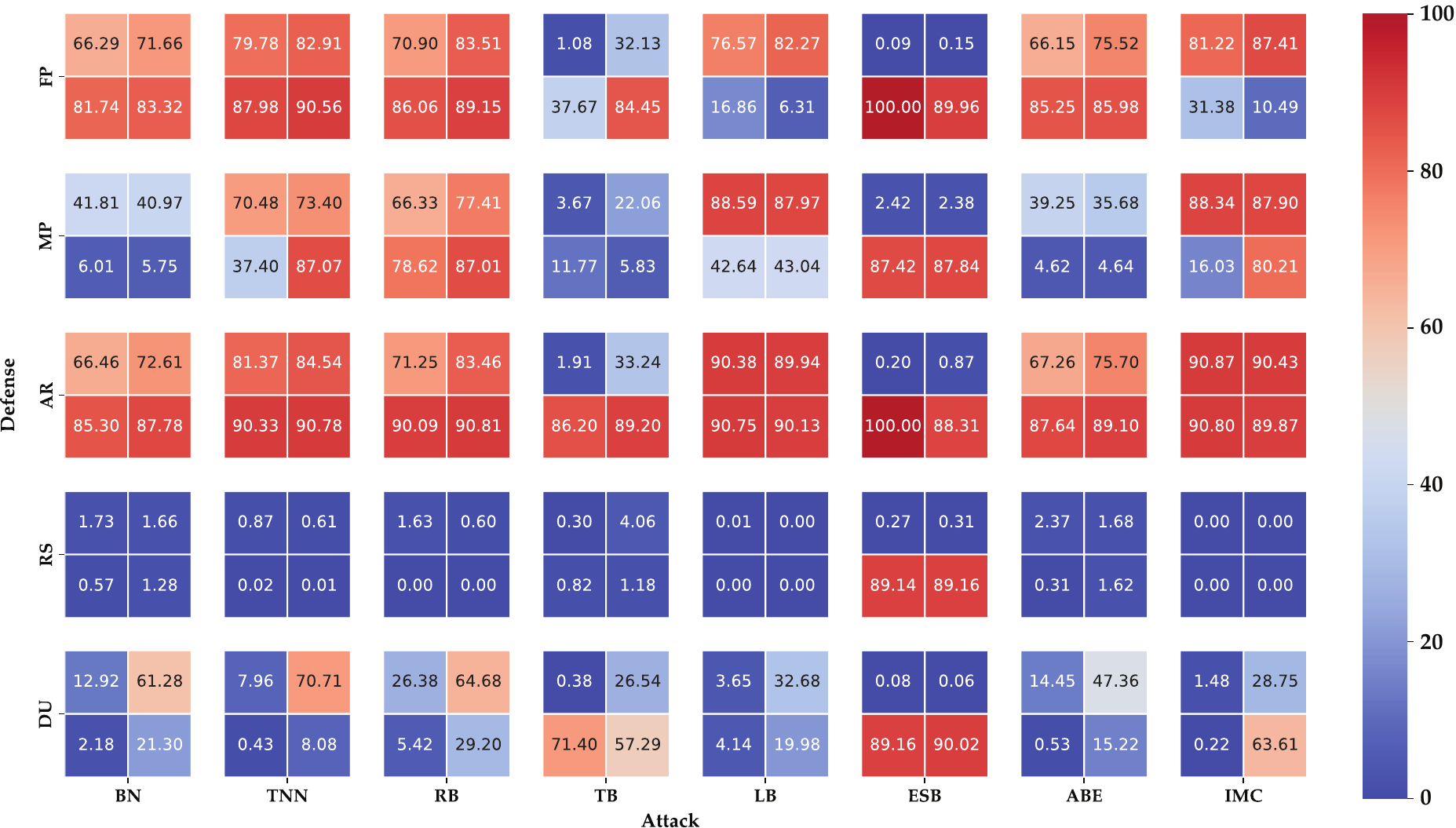}
    \caption{Impact of trigger definition on attack-agnostic defenses (left:  $|m|=3\times3$, right:  $|m|=6\times6$; lower:  $\alpha=0.0$, upper:  $\alpha=0.8$).}
    \label{fig:defense-trigger-general}
\end{figure}

Figure\mref{fig:defense-model-general} and\mref{fig:defense-trigger-general} shows the influence of DNN architecture and  trigger definition on the performance of attack-agnostic defenses (\mmp, \at, \rands, \du).

\begin{figure}[!ht]
    \centering
    \includegraphics[width=85mm]{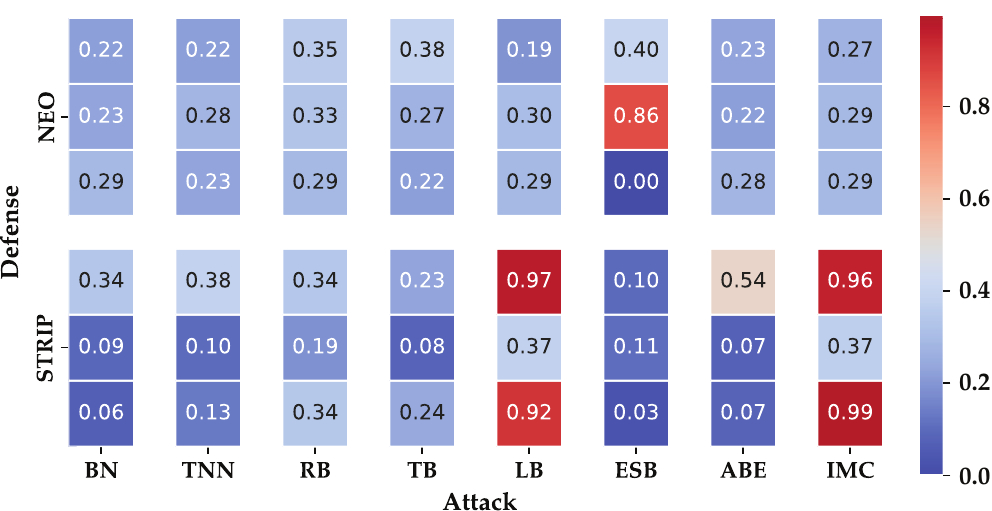}
    \caption{Impact of DNN architecture on input filtering defenses (lower:  ResNet18, middle:  DenseNet121; upper:  VGG13).}
    \label{fig:defense-model-inference}
\end{figure}

Figure\mref{fig:defense-model-inference} illustrate the impact of DNN architecture on the performance of input filtering defenses (\neo, \strip), which complements Figure\mref{fig:inference-trigger}.

\begin{figure}[!ht]
    \centering
    \includegraphics[width=85mm]{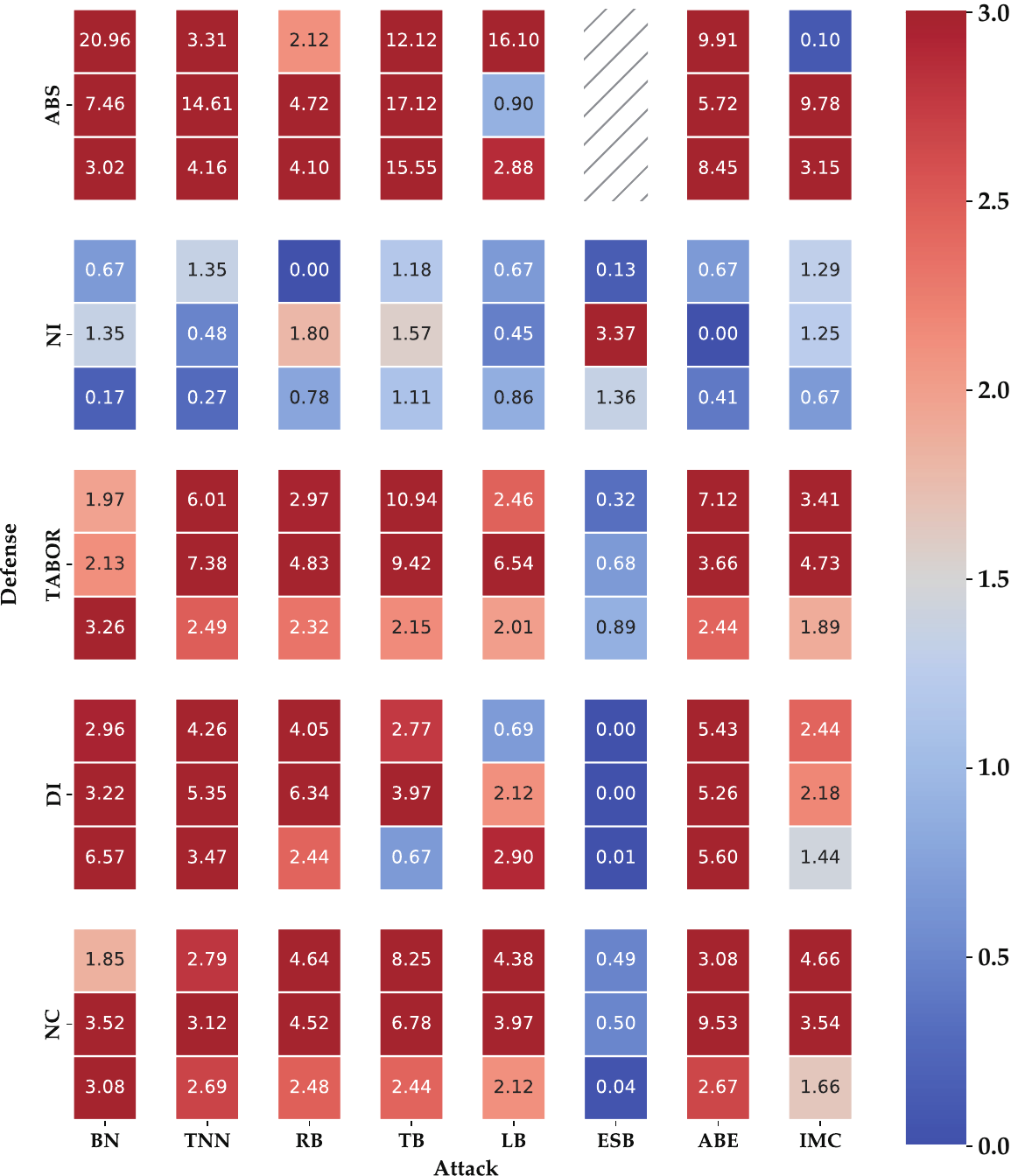}
    \caption{Impact of DNN architecture on model filtering defenses (lower:  ResNet18, middle:  DenseNet121; upper:  VGG13; note: \esb--\abs pair is inapplicable).}
    \label{fig:defense-model-model_inspection}
\end{figure}

\begin{figure}[!ht]
    \centering
    \includegraphics[width=90mm]{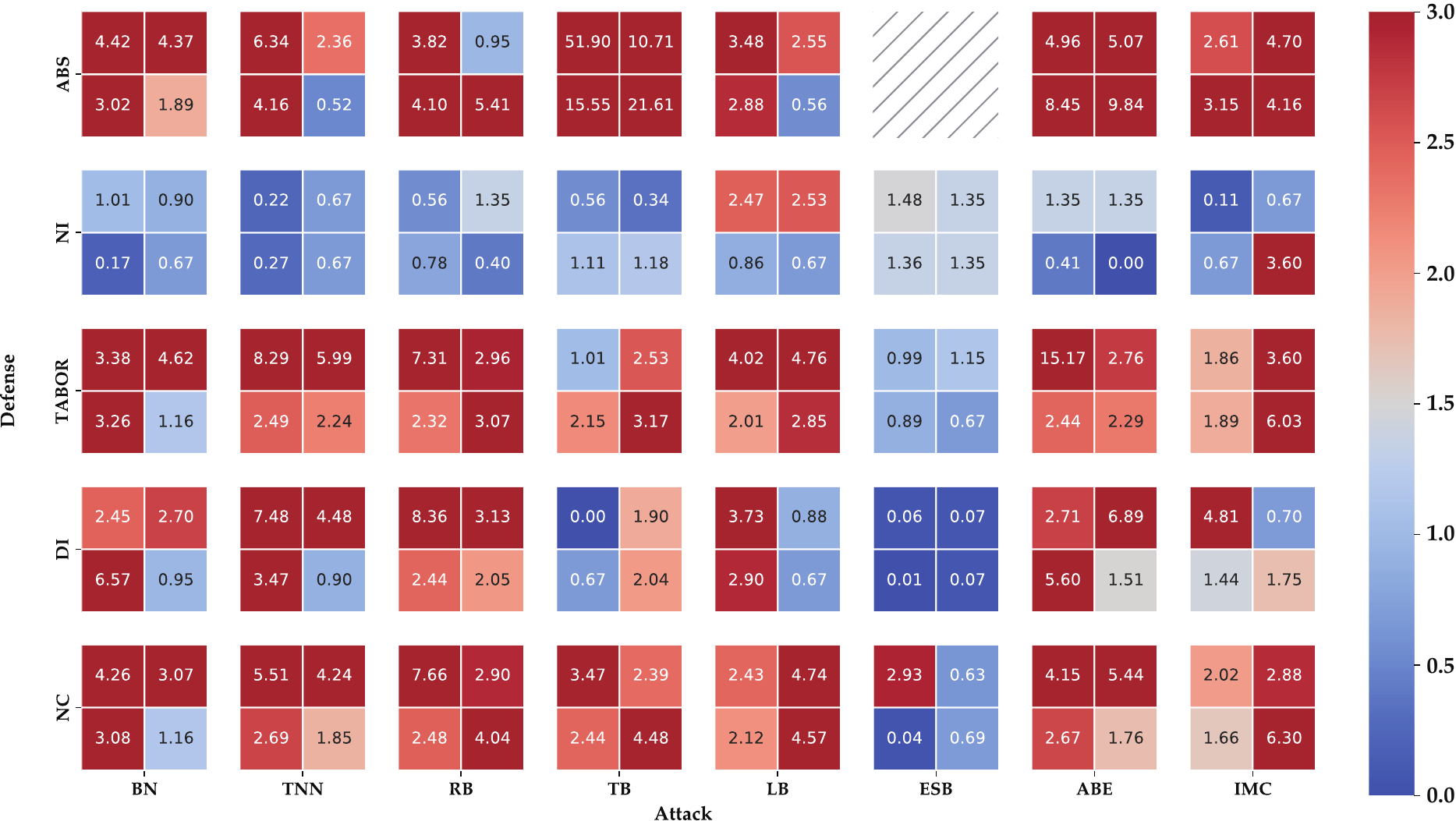}
    \caption{Impact of trigger definition on model filtering defenses (left:  $|m|=3\times3$, right:  $|m|=6\times6$; lower:  $\alpha=0.0$, upper:  $\alpha=0.8$; note: \esb--\abs pair is inapplicable).}
    \label{fig:defense-trigger-model_inspection}
\end{figure}

Figure\mref{fig:defense-model-model_inspection} and\mref{fig:defense-trigger-model_inspection} illustrate the impact of DNN architecture and  trigger definition on the performance of model-inspection defenses (\abs, \ninspect, \tabor, \di, \nc).

\end{document}